\documentclass[review]{elsarticle}

\usepackage{color}
\usepackage{graphicx}
\usepackage{tabularx}
\usepackage{multirow}
\usepackage{epstopdf}
\usepackage{amsmath}
\usepackage{amssymb}
\usepackage{bbding}
\usepackage{url}

\newcolumntype{?}{!{\vrule width 1pt}}

\journal{Image and Vision Computing}

\begin{document}

\begin{frontmatter}

\title{Unsupervised learning-based long-term superpixel tracking}

\author[l1,l2]{Pierre-Henri Conze \corref{cor1}}
\author[l3]{Florian Tilquin}
\author[l2,l4]{Mathieu Lamard}
\author[l3]{Fabrice Heitz}
\author[l2]{Gwenol\'e Quellec}

\address[l1]{IMT Atlantique, Technop\^ole Brest-Iroise, 29238 Brest, France}

\address[l2]{LaTIM UMR 1101, Inserm, 22 avenue Camille Desmoulins, 29238 Brest, France}

\address[l3]{ICube UMR 7357, CNRS, Universit\'e de Strasbourg, 300 boulevard S\'ebastien Brant, 67412 Illkirch, France}

\address[l4]{Universit\'e de Bretagne Occidentale, 2 avenue Foch, 29609 Brest, France \vspace{-0.5cm}}

\cortext[cor1]{corresponding author, \url{pierre-henri.conze@imt-atlantique.fr}}

\begin{abstract}
Finding correspondences between structural entities decomposing images is of high interest for computer vision applications. In particular, we analyze how to accurately track superpixels - visual primitives generated by aggregating adjacent pixels sharing similar characteristics - over extended time periods relying on unsupervised learning and temporal integration. A two-step video processing pipeline dedicated to long-term superpixel tracking is proposed. First, unsupervised learning-based superpixel matching provides correspondences between consecutive and distant frames using new context-rich features extended from greyscale to multi-channel and forward-backward consistency contraints. Resulting elementary matches are then combined along multi-step paths running through the whole sequence with various inter-frame distances. This produces a large set of candidate long-term superpixel pairings upon which majority voting is performed. Video object tracking experiments demonstrate the accuracy of our elementary estimator against state-of-the-art methods and proves the ability of multi-step integration to provide accurate long-term superpixel matches compared to usual direct and sequential integration. 
\vspace{-0.35cm} \\
\end{abstract}

\end{frontmatter}

\section{Introduction}
\label{sec:sec1}

Finding correspondences between multiple images is a fundamental problem in computer vision tasks including scene segmentation \cite{lezama2011track}, 3D reconstruction \cite{seitz2006comparison}, visual tracking \cite{yang2014robust}, trajectory analysis \cite{wang2013dense} or video editing like 2D-to-3D video conversion \cite{cao2011semi} or graphic elements propagation \cite{conze2016multi}. Established via local or global search, correspondences are usually either for the dense pixel grid as with optical flow \cite{farneback2003two} or sparse through key feature points \cite{shi1994good}. Alternatively, finding associations between structural entities decomposing images by grouping pixels enables semi-dense coverage of the whole image while drastically reducing the cost of correspondence. Matching a limited number of structural elements can also solve challenging issues such as large displacement, occlusions, appearance or illuminations changes. In this context, patch-based approximate nearest neighbor (ANN) search methods such as PatchMatch (PM) \cite{barnes2009patchmatch} and its extension to multi-resolution \cite{barnes2010generalized} are mainly used to find correspondences between patches. However, a regular decomposition of the image grid does not respect both object and motion contours and does not offer enough consistent support regions for image processing methods.

Contrary to image patches, superpixels - visual primitives generated by aggregating adjacent pixels sharing similar characteristics into semantic areas \cite{achanta2012slic} - offer more reliable support regions while preserving image geometry and object contours. Moreover, the hypothesis that motion discontinuities are a subset of photometric contours is usually used to preserve boundaries between objects exhibiting different motion. These findings have motivated recent optical flow algorithms using image data and smoothness terms adapted to the superpixel level \cite{chang2013superpixel,donne2015fast}. Conversely, we claim that image matching relying on superpixels could benefit from these advantages to offer more consistent associations than pixel or patch matches while providing a better management of motion discontinuities. In particular, this paper focuses on how to accurately find correspondences between superpixels over extended time periods.

Additionally to dense pixel-wise matching starting from superpixel-level pairings \cite{dong2016hsp2p}, superpixel correspondences have been already employed for visual tracking through superpixel-based discriminative appearance models \cite{yang2014robust} or object-background confidence maps \cite{fan2016visual}. However, these works perform superpixel matching based on comparisons of intrinsic superpixel features only, without taking full advantage of neighborhood information. Conversely, \cite{giraud2017superpatchmatch} exploits a structure of superpixel neighborhood called SuperPatch involved in a superpixel PM framework. Superpixel neighborhood information greatly improves correspondences since it alleviates some matching failures due to irregular decomposition of the same image content, not directly comparable between images. However, even with incorporated neighborhood information, directly computing a matching distance between irregular structures can be tedious, especially when images are divided into a large collection of superpixels.

A prior pixel-to-superpixel mapping can drive the matching at the superpixel level to provide more precise correspondences. In this direction, \cite{kanavati2017supervoxel} uses random forests (RF) \cite{breiman2001random} to establish supervoxel correspondences between two $3$D images in an unsupervised fashion. RF is trained on one image by using supervoxel indexes as voxel-wise class labels and robust context-rich features to describe the extended neighborhood. Applying RF on the other image yields a voxel labelling which is then regularised using majority voting within supervoxel boundaries. First validated on medical image registration, we explore the use of such learning-based superpixel matching for accurate superpixel matching across long video sequences.

Despite recent advances related to optical flow integration \cite{crivelli2015robust,conze2016multi}, the temporal tracking of superpixels over long-term video sequences has received little attention in the literature. \cite{wang2017constrained} uses a constrained graph where nodes denote superpixels and edges encode spatial, temporal, and appearance constraints. However, temporal constraints only model short-term smoothness between consecutive frames. The same finding arises in \cite{yang2014robust} whose tracker is conducted sequentially and therefore prone to motion drift. Establishing long-term superpixel correspondences requires to perform superpixel matching between consecutive and distant frames and therefore to handle simultaneously small and large displacements. To address this challenge, we exploit the concept of multi-step integration introduced for long-term motion estimation using optical flow \cite{conze2016multi}. The idea is to generate a large set of elementary displacement estimations performed between consecutive frames or with larger inter-frame distances. Once combined, elementary multi-step estimations result in a large set of long-term correspondences which are significative enough to be fused through statistical processing. We are not aware of any studies that have recovered this concept for long-term superpixel tracking while it could bring many benefits in our context. Indeed, it can alleviate matching errors during superpixel trajectory estimation since new steps can give a chance to match with a correct location again compared to sequential processing whose tracks may be lost. Moreover, statistical processing upon large representative long-term superpixel candidates can solve the uncertainty component present for matching tasks. 

It must be reported that deep learning has become popular for object tracking relying on convolutional networks to learn discriminative features to encode the target appearance \cite{ma2015hierarchical,li2016deeptrack} or recurrent networks trained with reinforcement learning to learn how to predict object locations across videos \cite{zhang2017deep}. Despite their high performance, these methods only provide very sparse bounding box tracking and do not describe how the boundaries of an irregular shaped object evolves in time as expected through long-term superpixel tracking.

In summary, two main contributions are proposed towards accurate long-term superpixel tracking. First, unsupervised learning-based superpixel matching is generalized and adapted from medical image processing \cite{kanavati2017supervoxel} to computer vision in order to find associations along sequences between consecutive and distant images decomposed into SLIC superpixels \cite{achanta2012slic} (Sect.\ref{sec:sec2}). The approach is carried out using classifiers such as $k$-nearest neighbors ($k$NN) or RF \cite{breiman2001random}, incorporates new forward-backward consistency contraints and fully exploits dedicated context-rich features we extended from greyscale \cite{glocker2014robust,kanavati2017supervoxel,conze2017semi} to multi-channel to incorporate neighborhood information on RGB frames. Second, based on this learning-based matching approach used as an elementary displacement estimator, we propose a multi-step integration strategy for long-term superpixel tracking. It combines multiple elementary superpixel matches obtained for some intermediate images following randomly selected multi-step paths (Sect.\ref{sec:sec3}). This produces a large set of candidate long-term superpixel pairings upon which a majority voting selection is performed. Based on object tracking experiments, Sect.\ref{sec:sec4} assesses the accuracy of the proposed elementary estimator against state-of-the-art methods and proves the ability of multi-step integration to provide an efficient long-term superpixel tracking compared to standard direct and sequential integration. We end with conclusions and perspectives in Sect.\ref{sec:sec5}.

\section{Superpixel matching with unsupervised learning}
\label{sec:sec2}

\footnotetext[1]{stands for \textit{first} and \textit{second}}

Let $\mathcal{V}$ be a video sequence of RGB images. Unsupervised learning-based superpixel matching is addressed between two consecutive or distant images $I_{f}$ and $I_{s}$ of $\mathcal{V}$. Each image $I_{q}:\Omega_{q}\subset\mathbb{N}^{2}\rightarrow\mathbb{N}^{3}$ associates a RGB color vector $I_{q}(\textbf{\textit{x}}_{q})$ to each pixel $\textbf{\textit{p}}_{q}$ located at $\textbf{\textit{x}}_{q} \in \Omega_{q}$ with $q\hspace{0.1cm}\in\hspace{0.1cm}\{f,s\}$\footnotemark[1].

\subsection{Problem formulation}
\label{ssec:sec2-1}

Let $\mathcal{F}=\{\textbf{\textit{f}}_{i}\}_{i\in\{1,\ldots,|\mathcal{F}|\}}$ and $\mathcal{S}=\{\textbf{\textit{s}}_{j}\}_{j\in\{1,\ldots,|\mathcal{S}|\}}$ be respectively the set of $|\mathcal{F}|$ and $|\mathcal{S}|$ connected superpixels partitioning $I_{f}$ and $I_{s}$. The superpixel decomposition is performed using the Simple Linear Iterative Clustering (SLIC) algorithm \cite{achanta2012slic} which aggregates neighboring pixels $\textbf{\textit{p}}_{q}$ based on spatial and intensity proximity criteria. Forward superpixel matching from $I_{f}$ to $I_{s}$  ($f<s$) consists in automatically learning a matching function $h_{f,s}$ that maps each superpixel $\textbf{\textit{f}}_{i}\in\mathcal{F}$ of $I_{f}$ to a given superpixel $\textbf{\textit{s}}_{j}\in\mathcal{S}$ of $I_{s}$ \cite{kanavati2017supervoxel} such that: 

\begin{equation}
\forall i \in \{1,\ldots,|\mathcal{F}|\}, \exists j\in\{1,\ldots,|\mathcal{S}|\} \hspace{0.1cm} | \hspace{0.1cm} h_{f,s}(\textbf{\textit{f}}_{i})=\textbf{\textit{s}}_{j}
\label{eq::sec2-1-eq1}
\end{equation}

\noindent Backward matching from $I_{s}$ to $I_{f}$ can be similarly considered by estimating $h_{s,f}$ mapping each superpixel $\textbf{\textit{s}}_{j}\in\mathcal{S}$ to a given superpixel $\textbf{\textit{f}}_{i}\in\mathcal{F}$. In what follows, learning-based superpixel matching is described in forward from $I_{f}$ to $I_{s}$.

\begin{figure}
\centering \begin{tabular}{cccc}
\hspace{-0.4cm} \includegraphics[width=2.7cm]{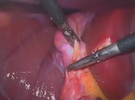} &
\hspace{-0.45cm} \includegraphics[width=2.7cm]{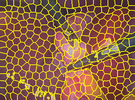} &
\hspace{-0.45cm} \includegraphics[width=2.7cm]{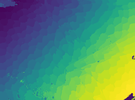} &
\hspace{-0.45cm} \includegraphics[width=2.7cm]{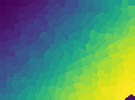} \vspace{-0.2cm} \cr
\hspace{-0.4cm} \footnotesize source $I_{f}$ &
\hspace{-0.45cm} \footnotesize SLIC \cite{achanta2012slic} &
\hspace{-0.45cm} \footnotesize pixel-wise prediction &
\hspace{-0.45cm} \footnotesize majority voting \vspace{0.1cm} \cr
\hspace{-0.4cm} \includegraphics[width=2.7cm]{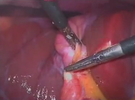} &
\hspace{-0.5cm} \includegraphics[width=2.7cm]{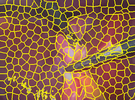} &
\hspace{-0.45cm} \includegraphics[width=2.7cm]{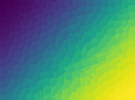} &
\hspace{-0.45cm} \includegraphics[width=2.4cm]{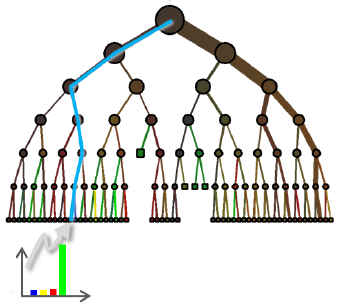} \vspace{-0.2cm} \cr
\hspace{-0.4cm} \footnotesize source $I_{s}$ &
\hspace{-0.45cm} \footnotesize SLIC \cite{achanta2012slic} &
\hspace{-0.45cm} \footnotesize pixel-wise training &
\hspace{-0.45cm} \footnotesize $k$NN or RF \cite{breiman2001random} \vspace{0.1cm} \cr
\end{tabular}
\caption{Superpixel matching between $I_{f}$ and $I_{s}$ using unsupervised learning applied with SLIC \cite{achanta2012slic} superpixel indexes as label entities followed by majority voting following \cite{kanavati2017supervoxel}. This example is produced for image pair $\{I_{50},I_{70}\}$ of the \texttt{lapa} sequence \cite{sznitman2012data} using RF \cite{breiman2001random}.}
\label{fig::sec2-2-fig-1} \vspace{-0.2cm}
\end{figure}

\subsection{Overall strategy}
\label{ssec:sec2-2}

Instead of relying on nearest neighbor search at the superpixel level through superpixel feature comparisons \cite{dong2016hsp2p,fan2016visual}, which is prone to ambiguity due to possible severe overlaps in feature space, we explore the use of pixel-wise $k$-nearest neighbors ($k$NN) or random forests (RF) \cite{breiman2001random} to establish correspondences between superpixels over-segmenting $I_{f}$ and $I_{s}$, as formulated in Eq.\ref{eq::sec2-1-eq1}. Usually employed with success for multi-class classification or regression, we show that such classifiers can also be used with profit for accurate superpixel correspondences. The overall learning-based superpixel matching strategy, illustrated Fig.\ref{fig::sec2-2-fig-1}, is carried out in an unsupervised manner. To give a powerful representation of global context, RF or $k$NN is considered with new pixel-wise context-rich features extended from greyscale \cite{glocker2014robust,kanavati2017supervoxel,conze2017semi} to multi-channel RGB and described Sect.\ref{ssec:sec2-3}.

The key idea is to perform training on the target image ($I_{s}$) by using superpixel indexes as pixel-wise class labels and testing on the source image ($I_{f}$) to get a pixel-to-superpixel mapping, as introduced in \cite{kanavati2017supervoxel}. In particular, the classifier aims at assigning a superpixel $\textbf{\textit{s}}_{j} \hspace{-0.08cm}\in \hspace{-0.08cm} \mathcal{S}$ to each pixel $\textbf{\textit{p}}_{f} \in \Omega_{f}$. A training set is thus built by considering all pixels $\textbf{\textit{p}}_{s}$ of $\Omega_{s}$ with their associated superpixel index, i.e. the index of the superpixel $\textbf{\textit{s}}_{j}$ they belong to. Once trained, the classifier is applied to $I_{f}$ to predict for each $\textbf{\textit{p}}_{f}$ the index of a superpixel of $\mathcal{S}$. This pixel-to-superpixel mapping is addressed in detail Sect.\ref{ssec:sec2-4}.

Mapping results are further regularized following superpixel boundaries to reach robust superpixel matches. Within each superpixel $\textbf{\textit{f}}_{i}$ of $I_{f}$, the most represented superpixel index among all pixel-wise predictions indicates the best superpixel match $h_{f,s}(\textbf{\textit{f}}_{i})$. This final superpixel matching step is detailed Sect.\ref{ssec:sec2-5} with new foward-backward (FW-BW) consistency constraints.

\begin{figure}
\centering \includegraphics[width=5.7cm]{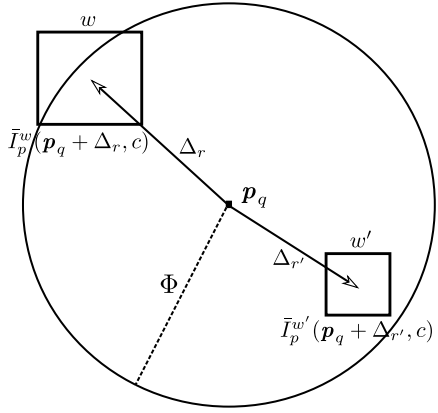} 
\vspace{-0.1cm}
\caption{Pixel-wise context-rich multi-channel features provide a description of the extended spatial context (see Sect.\ref{ssec:sec2-3} for further details).}
\label{fig::sec2-3-fig-1}
\end{figure}

\subsection{Context-rich multi-channel features}
\label{ssec:sec2-3}

Let $\bar{I}_{q}^{w}(\textbf{\textit{p}}_{q},c)$ be the average intensity on a local box of size $w$ centered on $\textbf{\textit{p}}_{q}$ located at $\textbf{\textit{x}}_{q}$ for channel $c\hspace{0.06cm}\in\hspace{0.06cm}\{r,g,b\}$. Pixel-wise context-rich features $\theta(\textbf{\textit{p}}_{q})=$ $\{\theta_{m}(\textbf{\textit{p}}_{q})\}_{m\in\{0,\ldots,K_{a}-1\}}$ assigned to pixels $\textbf{\textit{p}}_{q}$ are extended from greyscale \cite{glocker2014robust,kanavati2017supervoxel,conze2017semi} to multi-channel as follows: \vspace{-0.4cm}

\begin{eqnarray}
\hspace{-0.2cm} \theta_{m}(\textbf{\textit{p}}_{q}) \hspace{-0.005cm} = \hspace{-0.005cm} \bar{I}_{p}^{w}(\textbf{\textit{p}}_{q}+\Delta_r,c) - \beta \times \bar{I}_{p}^{w'}(\textbf{\textit{p}}_{q}+\Delta_{r'},c)
\label{eq::sec2-3-eq1}
\end{eqnarray}

\noindent where displacements $\Delta_{\{r,r'\}}$ are randomly defined starting from $\textbf{\textit{p}}_{q}$ in a disc of maximal radius $\Phi$ (Fig.\ref{fig::sec2-3-fig-1}). $\beta\hspace{-0.075cm}\in\hspace{-0.075cm}\{0,1\}$ is a binary parameter which focuses whether on intensity differences between two boxes randomly located in the extended neighborhood ($\beta\hspace{-0.025cm}=\hspace{-0.025cm}1$) or on the value obtained from one single box only ($\beta\hspace{-0.025cm}=\hspace{-0.025cm}0$). Color intensities around $\textbf{\textit{p}}_{q}$ are included in the feature vector $\theta(\textbf{\textit{p}}_{q})$ by forcing $\Delta_r=\beta=0$ for all possible pre-defined box sizes $w$.

By randomly generating many different box sizes $w$ and offsets $\Delta_r$, we obtain a large set of $K_{a}$ features describing the extended spatial context for all color channels $c$. Parameters $\{w,w',\Delta_r,\Delta_r',\textcolor{black}{\beta},c\}$ are randomly generated once and remain similar whatever the image $I_{f}$ or $I_{s}$ under consideration.

\subsection{Pixel-to-superpixel mapping}
\label{ssec:sec2-4}

Pixel-to-superpixel mapping relies on machine learning to compute pixel-to-superpixel mapping probabilities denoted as $\texttt{p}(h_{f,s}(\textbf{\textit{p}}_{f}) \hspace{-0.1cm}= \hspace{-0.1cm} \textbf{\textit{s}}_{n})$ for each pixel $\textbf{\textit{p}}_{f} \in \Omega_{f}$ with respect to all superpixels $\textbf{\textit{s}}_{n} \hspace{-0.03cm} \in \hspace{-0.01cm} \mathcal{S}$ with $n\in\{1,\ldots,|\mathcal{S}|\}$. 

The procedure with random forests (RF) \cite{breiman2001random} is conducted as follows. The forest is formed by $T$ uncorrelated trees made of both internal nodes splitting data according to binary tests $\Psi$ and leaf nodes which reach all together a final data partition. At each internal node, the split sends pixels $\textbf{\textit{p}}_{q}$ to left and right child nodes during training ($q\hspace{-0.05cm}=\hspace{-0.075cm}f$) and prediction ($q\hspace{-0.05cm}=\hspace{-0.075cm}s$). The associated binary test $\Psi$ focuses on a random subset $\hat{\theta}(\textbf{\textit{p}}_{q})$ of context-rich multi-channel visual features $\theta(\textbf{\textit{p}}_{q})$ assigned to $\textbf{\textit{p}}_{q}$ (Sect.\ref{ssec:sec2-4}) and divides the input pixel set based on the following split rule: \vspace{-0.5cm} \\

\begin{equation}
\Psi(\textbf{\textit{p}}_{q},\theta(\textbf{\textit{p}}_{q}))= 
\left\{
\begin{array}{l}
\textrm{true, if } \hat{\theta}(\textbf{\textit{p}}_{q})>\tau \\
\textrm{false, otherwise} \\
\end{array}
\right.
\label{eq::sec2-4-eq1}
\end{equation}

\noindent where $\hat{\theta}(\textbf{\textit{p}}_{q})$ is compared to a threshold $\tau$ and $q\in\{f,s\}$.

Internal node parameters ($\{\tau,\hat{\theta}(\textbf{\textit{p}}_{q})\}$) are optimized \textit{via} information gain maximization with respect to the training dataset $\mathcal{L}=\{\textbf{\textit{p}}_{s}, c(\textbf{\textit{s}}_{j})\}$ combining pixels $\textbf{\textit{p}}_{s}$ belonging to $\textbf{\textit{s}}_{j}$ with their associated superpixel index $c(\textbf{\textit{s}}_{j})=j$ with $j \hspace{-0.05cm}  \in \hspace{-0.05cm}   \{1,\ldots,|\mathcal{S}|\}$. After optimization, each leaf node $l_{t}$ of the $t^{\text{th}}$ tree receives a partition $\mathcal{L}_{l_{t}}$ of $\mathcal{L}$ and produces the class probability distribution $\texttt{p}_{l_{t}}(c(\textbf{\textit{s}}_{j})|\mathcal{L})$ for all superpixels $\textbf{\textit{s}}_{j}$. 

To predict the corresponding superpixel index $c(\textbf{\textit{s}}_{n})$ of a given pixel $\textbf{\textit{p}}_{f}\hspace{-0.075cm}\in\hspace{-0.015cm}\Omega_{f}$ with associated  visual features $\theta(\textbf{\textit{p}}_{f})$ during testing, $\textbf{\textit{p}}_{f}$ is injected into each optimized tree and finally reaches a leaf node $l_{t}$ per tree following the successive split rules (Eq.\ref{eq::sec2-4-eq1}). The pixel-to-superpixel mapping probability $\texttt{p}(h_{f,s}(\textbf{\textit{p}}_{f}) = \textbf{\textit{s}}_{n})$ denoting the probability that $\textbf{\textit{s}}_{n}$ is assigned to $\textbf{\textit{p}}_{f}$ is obtained for each $\textbf{\textit{s}}_{n}$ by: \vspace{-0.4cm}

\begin{eqnarray}
\texttt{p}(h_{f,s}(\textbf{\textit{p}}_{f}) = \textbf{\textit{s}}_{n}) =
\frac{1}{T} \sum_{t=1}^{T} \texttt{p}_{l_{t}}(c(\textbf{\textit{s}}_{n})|\mathcal{L})
=  \frac{1}{T} \sum_{t=1}^{T} 
\frac{|\{\textbf{\textit{p}}_{s},c(s_{l})\}\in \mathcal{L}_{l_{t}} \text{ }|\text{ } l=n \hspace{0.05cm} |}{|\mathcal{L}_{l_{t}}|}
\label{eq::sec2-4-eq2}
\end{eqnarray}

Contrary to RF, the $k$NN classifier simply stores instances of training data insstead of building a general internal model. Pixel-to-superpixel mapping probabilities $\texttt{p}(h_{f,s}(\textbf{\textit{p}}_{f})\hspace{-0.1cm}=\hspace{-0.1cm}\textbf{\textit{s}}_{n})$ are computed by looking at the class (superpixel index) distribution among the $k$ nearest neighbors $\textbf{\textit{p}}_{s}$ of each pixel $\textbf{\textit{p}}_{f}$ in feature space. Nearest neighbors are estimated using Euclidean distance on context-rich features (Sect.\ref{ssec:sec2-3}) assigned to pixels $\textbf{\textit{p}}_{q}$ with $q\in\{f,s\}$.

\subsection{Superpixel-to-superpixel matching}
\label{ssec:sec2-5}

Once pixel-to-superpixel mapping probabilities $\texttt{p}(h_{f,s}(\textbf{\textit{p}}_{f}) \hspace{-0.2cm} = \hspace{-0.2cm} \textbf{\textit{s}}_{n})$ are computed for each superpixel $\textbf{\textit{s}}_{n} \hspace{-0.05cm} \in \hspace{-0.03cm} \mathcal{S}$ using context-rich features (Sect.\ref{ssec:sec2-3}) involved in RF or $k$NN (Sect.\ref{ssec:sec2-4}), two steps are required to get final superpixel pairings. First, the final pixel-to-superpixel mapping for each $\textbf{\textit{p}}_{f}$ of $I_{f}$ can be found using: 

\begin{equation}
h_{f,s}(\textbf{\textit{p}}_{f}) = \textbf{\textit{s}}_{j} = \arg\max_{\textbf{\textit{s}}_{n} \in \mathcal{S}} \texttt{p}(h_{f,s}(\textbf{\textit{p}}_{f}) = \textbf{\textit{s}}_{n})
\label{eq::sec2-5-eq1}
\end{equation} 

\noindent Second, majority voting among all pixels of a given superpixel $\textbf{\textit{f}}_{i} \in \mathcal{F}$ can be performed by selecting the most represented superpixel index. The final matching $h_{f,s}(\textbf{\textit{f}}_{i})=\textbf{\textit{s}}_{j}$ is defined such that $c(\textbf{\textit{s}}_{j})$ satisfies: \vspace{-0.5cm}

\begin{eqnarray}
c(\textbf{\textit{s}}_{j}) = \arg\max \texttt{hist}(\{c(h_{f,s}(\textbf{\textit{p}}_{f})) \hspace{0.1cm} | \hspace{0.1cm} \textbf{\textit{p}}_{f} \in \textbf{\textit{f}}_{i} \})
\label{eq::sec2-5-eq2}
\end{eqnarray} 

\noindent An alternative consists in averaging the pixel-to-superpixel mapping probabilities at the superpixel level instead of making hard decision for each $\textbf{\textit{p}}_{f}$ as performed in Eq.\ref{eq::sec2-5-eq1}: \vspace{-0.2cm}

\begin{equation}
\texttt{p}(h_{f,s}(\textbf{\textit{f}}_{i}) = \textbf{\textit{s}}_{n}) = \frac{1}{|\textbf{\textit{f}}_{i}|} \displaystyle \sum_{ \textbf{\textit{p}}_{f}\in \text{ } \textbf{\textit{f}}_{i}}  \texttt{p}(h_{f,s}(\textbf{\textit{p}}_{f}) = \textbf{\textit{s}}_{n})
\label{eq::sec2-5-eq3}
\end{equation} 

\noindent We keep at this point all possible outcomes between candidate matches. Decisions are postponed to the superpixel level by finding the superpixel $\textbf{\textit{s}}_{j} \in \mathcal{S}$ which maximizes $\texttt{p}(h_{f,s}(\textbf{\textit{f}}_{i}) = \textbf{\textit{s}}_{n})$: \vspace{-0.7cm} \\

\begin{equation}
h_{f,s}(\textbf{\textit{f}}_{i}) = \textbf{\textit{s}}_{j} = \arg\max_{\textbf{\textit{s}}_{n} \in \mathcal{S}} \texttt{p}(h_{f,s}(\textbf{\textit{f}}_{i}) = \textbf{\textit{s}}_{n})
\label{eq::sec2-5-eq4}
\end{equation} 

Forward-backward consistency can be enforced in the context where two mapping functions are learned: $h_{f,s}$ (resp. $h_{s,f}$) that maps each supervoxel $\textbf{\textit{f}}_{i}\hspace{-0.08cm}\in\hspace{-0.05cm}\mathcal{F}$ ($\textbf{\textit{s}}_{j}\hspace{-0.08cm}\in\hspace{-0.05cm}\mathcal{S}$) to a given $\textbf{\textit{s}}_{j}$ ($\textbf{\textit{f}}_{i}$) belonging to $\mathcal{S}$ ($\mathcal{F}$) in forward (backward). Thus, we extend Eq.\ref{eq::sec2-5-eq4} with a new consistency constraint that guides the mutual matching between $\textbf{\textit{f}}_{i}$ and $\textbf{\textit{s}}_{j}$: \vspace{-0.7cm} \\

\begin{equation}
\hspace{-0.2cm} h_{f,s}(\textbf{\textit{f}}_{i}) = \arg\max_{\textbf{\textit{s}}_{n} \in \mathcal{S}} \text{ } \texttt{p}(h_{f,s}(\textbf{\textit{f}}_{i}) \hspace{-0.05cm}= \hspace{-0.05cm}\textbf{\textit{s}}_{n}) \times \texttt{p}(h_{s,f}(\textbf{\textit{s}}_{n}) \hspace{-0.05cm}=\hspace{-0.05cm} \textbf{\textit{f}}_{i})
\label{eq::sec2-5-eq5}
\end{equation} 

The whole unsupervised learning-based strategy described above can be performed all along the video $\mathcal{V}$ to match superpixels decomposing consecutive or distant images, both in forward and backward directions.

\section{Long-term superpixel tracking using multi-step integration}
\label{sec:sec3}

We address at this stage long-term superpixel tracking for sequence $\mathcal{V}$ composed of $N\hspace{-0.05cm}+\hspace{-0.05cm}1$ RGB frames $I_{n}:\Omega_{n}\subset\mathbb{N}^{2}\rightarrow\mathbb{N}^{3}$ using the learning-based superpixel matching strategy, described Sect.\ref{sec:sec2} for a given pair of consecutive or distant frames, as elementary estimator. Each frame $I_{n}$ is decomposed into a set of superpixels obtained using SLIC \cite{achanta2012slic} with the same compactness parameter, i.e. same weighting between spatial and intensity proximity. One particular frame (usually the first one) of $\mathcal{V}$ is defined as the reference frame and denoted $I_{ref}$. In this context, we aim at finding correspondences between superpixels over-segmenting $I_{ref}$ and superpixels defined in frames $I_{n}$ with $n\hspace{-0.05cm}\in\hspace{-0.05cm}\{0,\ldots,N\}\hspace{-0.025cm}\neq \hspace{-0.025cm}ref$. Let $\mathcal{F}=\{\textbf{\textit{f}}_{i}\}_{i\in\{1,\ldots,|\mathcal{F}|\}}$ and $\mathcal{S}=\{\textbf{\textit{s}}_{j}\}_{j\in\{1,\ldots,|\mathcal{S}|\}}$ be respectively the set of $|\mathcal{F}|$ and $|\mathcal{S}|$ connected superpixels partitioning $I_{ref}$ and $I_{n}$.

Both superpixel trajectory estimation between the reference frame and all the images of the sequence and superpixel matching to match each image to the reference frame can be considered, as in \cite{crivelli2015robust,conze2016multi}. From-the-reference estimation is useful for information pushing from superpixels of $I_{ref}$ whereas to-the-reference estimation allows information propagation over superpixels of each frame $I_{n}$ by pulling it from $I_{ref}$. The description below focuses on a given pair $\{I_{ref},I_{n}\}$ where $I_{n}$ is located far away from $I_{ref}$. Correspondences for the whole sequence are obtained by processing each pair $\{I_{ref},I_{n}\}$ independently $\forall n \hspace{-0.06cm} \neq \hspace{-0.06cm} ref$.

Starting from learning-based superpixel matching (Sect.\ref{sec:sec2}) as elementary motion estimator, two temporal integration schemes can be considered at first glance to find $h_{ref,n}$ mapping each superpixel $\textbf{\textit{f}}_{i}\hspace{-0.05cm}\in\hspace{-0.05cm}\mathcal{F}$ to a given superpixel $\textbf{\textit{s}}_{j}$ of $\mathcal{S}$ such that $h_{ref,n}(\textbf{\textit{f}}_{i})=\textbf{\textit{s}}_{j}$. First, sequential integration can be employed passing through all intermediate frames, similarly to dense point tracking algorithms \cite{brox2011large}. This step-by-step strategy can gradually apprehend appearance changes and large displacements but it may lead to large error accumulation resulting in a substantial drift over extended time periods. This drawback is further enhanced when using superpixels since superpixel decompositions across the sequence may result in an irregular partitioning of the image content. Second, to avoid error accumulations, direct matching \cite{roth2009discrete} can be applied between superpixels of $I_{ref}$ and $I_{n}$, exactly as in Sect.\ref{sec:sec2}. However, this ignores that $\mathcal{V}$ consists of inter-related images with redundant and smoothly evolving content, which makes large displacement and aspect changes challenging to handle. 

Issues related to both sequential and direct superpixel tracking could be partially compensated by complexifying the superpixel matching models and criteria, but an uncertainty component remains. This argues in favor of a statistical processing (Sect.\ref{ssec:sec3-3}) which takes into account a large set of candidate long-term superpixel matches (Sect.\ref{ssec:sec3-2}) obtained using multi-step combination of elementary superpixel pairings previously established through unsupervised learning following Sect.\ref{sec:sec2}.

\subsection{Multi-step integration of elementary matches}
\label{ssec:sec3-2}

Multi-step integration aims at producing a large set of candidate long-term superpixel pairings between $I_{ref}$ and $I_{n}$ using intermediate superpixel correspondences to form a significative set of samples upon which statistical selection (Sect.\ref{ssec:sec3-3}) is relevant. Formely introduced for optical flow integration \cite{conze2016multi}, we show that this heuristic can be extended towards accurate long-term superpixel tracking. As inputs, we take a set of superpixel match fields pre-estimated from each frame of $\mathcal{V}$ including $I_{ref}$. These matches are computed between consecutive frames or with larger inter-frame distances \cite{conze2016multi} using learning-based superpixel matching (Sect.\ref{sec:sec2}). Let $\mathcal{A}_{n}\hspace{-0.02cm}=\hspace{-0.02cm}\{\alpha_{1},\alpha_{2},\ldots,\alpha_{Q_{n}}\}\hspace{-0.02cm}\subset\hspace{-0.02cm}{\{1,\ldots,N\hspace{-0.06cm}-\hspace{-0.06cm}n\}}$ be the set of $Q_{n}$ possible steps at instant $n$ which means that $\{h_{n,n+\alpha_{1}},h_{n,n+\alpha_{2}},\ldots,h_{n,n+\alpha_{Q_{n}}}\}$ has been previously learned using $k$NN or RF. Thus, for each step $\alpha_{q} \in \mathcal{A}_{n}$, we have a superpixel match in $I_{n+\alpha_{q}}$ for each superpixel of $I_{n}$ through the mapping function $h_{n,n+\alpha_{q}}$, and this for each frame.

\begin{figure}
\begin{center}
\includegraphics[height=4cm]{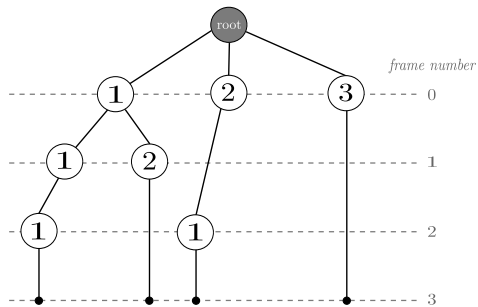} 
\end{center} 
\caption{Generation of step sequences from $I_{0}$ to $I_{3}$ with steps $1$, $2$, and $3$ by creating a tree structure: $\boldsymbol{\Gamma}_{0,3}=\{\{1,1,1\},\{1,2\},\{2,1\},\{3\}\}$.}
\label{fig:sec3-2-fig1}
\end{figure}

\begin{figure*}
\begin{center}
\includegraphics[height=2.9cm]{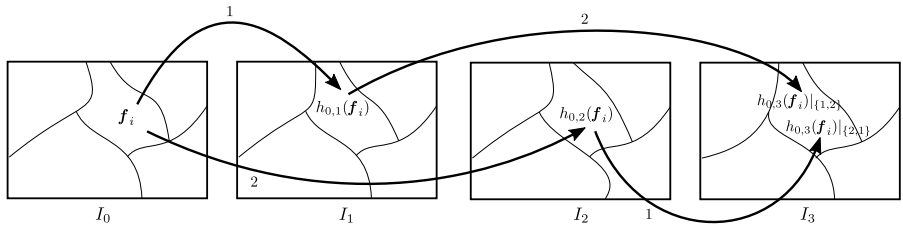} 
\end{center} 
\caption{Generation of multi-step paths corresponding to step sequences $\{\{1,2\},\{2,1\}\}\subset\boldsymbol{\Gamma}_{0,3}$ from $I_{0}$ to $I_{3}$.}
\label{fig:sec3-2-fig2}
\end{figure*}

The starting point of multi-step integration consists in initially generating all possible step sequences (Fig.\ref{fig:sec3-2-fig1}), i.e. combinations of steps, to join $I_{n}$ from $I_{ref}$. Then, each generated step sequence defines a multi-step path (Fig.\ref{fig:sec3-2-fig2}) linking each superpixel $\textbf{\textit{f}}_{i}$ of $I_{ref}$ to a superpixel $\textbf{\textit{s}}_{j}$ in $I_{n}$ passing through superpixels of some intermediate frames.

Let $\boldsymbol{\Gamma}_{ref,n}=\{\boldsymbol{\gamma}_{0},\boldsymbol{\gamma}_{1},\ldots,\boldsymbol{\gamma}_{K-1}\}$ be the set of the $K$ possible step sequences $\boldsymbol{\gamma}_{k}$ between $I_{ref}$ and $I_{n}$. A step sequence $\boldsymbol{\gamma}_{k}=\{\alpha_{1}^{k},\alpha_{2}^{k},\ldots,\alpha_{K_{\boldsymbol{\gamma}_{k}}}^{k}\}$ is defined by a set of $K_{\boldsymbol{\gamma}_{k}}$ steps which once cascaded join $I_{n}$ from $I_{ref}$. $\boldsymbol{\Gamma}_{ref,n}$ is computed by building a tree structure (Fig.\ref{fig:sec3-2-fig1}) where each node corresponds to a field of superpixel matches assigned to a given frame for a given step value (node value). Going from the root node to leaf nodes of this tree structure gives the possible step sequences which are stacked into $\boldsymbol{\Gamma}_{ref,n}$. For instance, the tree displayed Fig.\ref{fig:sec3-2-fig1} indicates the $4$ possible step sequences from $I_{0}$ to $I_{3}$ with steps $1$, $2$, and $3$: $\boldsymbol{\Gamma}_{0,3}=\{\{1,1,1\},\{1,2\},\{2,1\},\{3\}\}$.

Once all the $K$ possible step sequences $\boldsymbol{\gamma}_{k}$ between $I_{ref}$ and $I_{n}$ are generated, the corresponding multi-step paths are constructed (Fig.\ref{fig:sec3-2-fig2}). For step sequence $\boldsymbol{\gamma}_{k}=\{\alpha_{1}^{k},\alpha_{2}^{k},\ldots,\alpha_{K_{\boldsymbol{\gamma}_{k}}}^{k}\} \in \boldsymbol{\Gamma}_{ref,n}$ composed of $K_{\boldsymbol{\gamma}_{k}}$ steps, superpixel matching between $I_{ref}$ and $I_{n}$ is performed via: \vspace{-0.5cm} 

\begin{eqnarray}
\hspace{-0.5cm} \left. h_{ref,n}(\textbf{\textit{f}}_{i}) \right|_{\boldsymbol{\gamma}_{k}} \hspace{-0.5cm} &=& \hspace{-0.1cm}
h_{ref+\sum_{p=1}^{K_{\boldsymbol{\gamma}_{k}}-1}\alpha_{p}^{k},n} 
\circ \ldots \circ \nonumber \\
&& h_{ref+\alpha_{1}^{k},ref+\alpha_{1}^{k}+\alpha_{2}^{k}}
\circ h_{ref,ref+\alpha_{1}^{k}}(\textbf{\textit{f}}_{i})
\label{eq::sec3-2-eq1}
\end{eqnarray}

\noindent with $ref+\sum_{p=1}^{K_{\boldsymbol{\gamma}_{k}}}\alpha_{p}^{k}=n$. Once all the steps $\alpha_{j}^{k}\in{\boldsymbol{\gamma}_{k}}$ have been run through, one gets $\left. h_{ref,n}(\textbf{\textit{f}}_{i}) \right|_{\boldsymbol{\gamma}_{k}}$, the superpixel in $I_{n}$ corresponding to $\textbf{\textit{f}}_{i} \in I_{ref}$ obtained with step sequence $\boldsymbol{\gamma}_{k}$. For $\boldsymbol{\gamma}_{k}=\{1,2\}\ \in \boldsymbol{\Gamma}_{0,3}$ for instance (Fig.\ref{fig:sec3-2-fig2}), we have: \vspace{-0.4cm} 

\begin{eqnarray}
\hspace{-0.1cm} \left. h_{0,3}(\textbf{\textit{f}}_{i}) \right|_{\{1,2\}} =  h_{1,3}\big(h_{0,1}(\textbf{\textit{f}}_{i})\big) = h_{1,3} \circ h_{0,1}(\textbf{\textit{f}}_{i}) 
\label{eq::sec3-2-eq2}
\end{eqnarray}

A large set of candidate superpixels in $I_{n}$ is finally reached by considering all the step sequences of $\boldsymbol{\Gamma}_{ref,n}$ and this for each superpixel $\textbf{\textit{f}}_{i}$ defined in $I_{ref}$. Thus, to each $\textbf{\textit{f}}_{i}$ is associated a large set of candidate superpixels in $I_{n}$ defined as $\mathcal{T}_{ref,n}(\textbf{\textit{f}}_{i})=\{\left. h_{ref,n}(\textbf{\textit{f}}_{i}) \right|_{\boldsymbol{\gamma}_{k}}\}_{k\in\{0,1,\ldots,K\}}$.

Multi-step integration has been previously presented as an exhaustive candidate generation process. In practice, selecting only a subset of all possible step sequences and therefore associated multi-step paths is required to be able to build and keep in memory the multi-step integration stage outputs growing exponentially \cite{conze2016multi}. Overall, $5877241$ multi-step paths can be generated for a distance of $30$ frames with steps $1$, $2$, $5$ and $10$ for instance. Up to a few thousands can be actually considered to avoid computational and memory issues. The selection of $L$ step sequences $\boldsymbol{\Gamma}_{ref,n}^{*}\hspace{0.08cm}=\{\boldsymbol{\gamma}_{0},\boldsymbol{\gamma}_{1},\ldots,\boldsymbol{\gamma}_{L-1}\}$ among the $K$ possible step sequences $\boldsymbol{\Gamma}_{ref,n}$ is therefore necessary, with $L<<K$.

Two complexity reduction rules are taken from \cite{conze2016multi}. We start by removing the largest step sequences in terms of number of constituting steps. A threshold of $K_{max}$ number of steps is thus set and only step sequences $\boldsymbol{\gamma}_{k}=\{\alpha_{1}^{k},\alpha_{2}^{k},\ldots,\alpha_{K_{\boldsymbol{\gamma}_{k}}}^{k}\}$ for which $K_{\boldsymbol{\gamma}_{k}} \hspace{-0.08cm} \leq \hspace{-0.03cm} K_{max}$ are kept. Indeed, too many steps may induce an important drift due to multiple intermediates. Then, random selection of $N_{max}$ step sequences among remaining ones is performed.

\subsection{Long-term match selection}
\label{ssec:sec3-3}

Once step selection is performed, we obtain for each superpixel $\textbf{\textit{f}}_{i}$ of $I_{ref}$ a set of $L$ candidate superpixels $\mathcal{T}_{ref,n}(\textbf{\textit{f}}_{i})=\{\left. h_{ref,n}(\textbf{\textit{f}}_{i}) \right|_{\boldsymbol{\gamma}_{l}\in\boldsymbol{\Gamma}_{ref,n}^{*}}\}$ defined in $I_{n}$ with $l\hspace{-0.05cm}\in\hspace{-0.05cm}\{0,...,L\hspace{-0.01cm}-\hspace{-0.01cm}1\}$. The final candidate selection is performed via majority voting among $\mathcal{T}_{ref,n}(\textbf{\textit{f}}_{i})$, i.e. the final matching $h_{ref,n}(\textbf{\textit{f}}_{i}) \hspace{-0.06cm} = \hspace{-0.06cm} \textbf{\textit{s}}_{j}$ is defined such that $c(\textbf{\textit{s}}_{j})$ satisfies: \vspace{-0.5cm}

\begin{eqnarray}
\hspace{-0.6cm} c(\textbf{\textit{s}}_{j}) = \arg\max \texttt{hist}(\{c(\left. h_{ref,n}(\textbf{\textit{f}}_{i}) \right|_{\boldsymbol{\gamma}_{l}}) \hspace{0.1cm} | \hspace{0.1cm} \boldsymbol{\gamma}_{l} \in \boldsymbol{\Gamma}_{ref,n}^{*} \})
\label{eq::sec3-3-eq1}
\end{eqnarray} 

\noindent Thank to the random step sequence selection (Sect.\ref{ssec:sec3-2}), the set of generated superpixel candidates is both significative and uncorrelated enough to assume that the most represented superpixel provides an accurate superpixel match. 

Forward-backward consistency can be also considered in this context by providing to-the-reference multi-step paths additionally to from-the-reference ones. We thus incorporate in $\mathcal{T}_{ref,n}(\textbf{\textit{f}}_{i})$ superpixels $ \textbf{\textit{s}}_{j}$ such that $\left. h_{n,ref}(\textbf{\textit{s}}_{j})\right|_{\boldsymbol{\gamma}_{l}\in\boldsymbol{\Gamma}_{n,ref}^{*}}=\textbf{\textit{f}}_{i}$ where $\boldsymbol{\Gamma}_{n,ref}^{*}$ is the set of $L$ selected step sequences in the to-the-reference direction. The resulting additional superpixel candidates are referred as \textit{reverse} candidates in opposition to \textit{direct} ones, i.e. those which were formerly stacked into $\mathcal{T}_{ref,n}(\textbf{\textit{f}}_{i})$. To further guide mutual matching between $\textbf{\textit{f}}_{i}$ and $\textbf{\textit{s}}_{j}$, one can apply majority voting (Eq.\ref{eq::sec3-3-eq1}) only on superpixels generated in both from/to-the-reference directions. 

Superpixel correspondences with respect to $I_{ref}$ are provided for the whole sequence relying on multi-step integration applied independently for each pair $\{I_{ref},I_{n}\}$ $\forall n \neq ref$ and based on unsupervised learning-based superpixel matching as an elementary estimator.

\section{Application to video object tracking}
\label{sec:sec4}

Different aspects of the proposed methodology are evaluated through video object tracking experiments. First, the ability of unsupervised learning-based superpixel matching to provide a reliable accurate elementary estimator between consecutive and distant frames is proven with comparisons to state-of-the-art methods (Sect.\ref{ssec:sec4-1}). Second, the capacity of the proposed multi-step integration stage to perform robust long-term superpixel tracking is shown using both $k$NN and RF-based multi-step elementary superpixel matches (Sect.\ref{ssec:sec4-2}). Moreover, multi-step integration results are assessed with respect to straightforward direct and sequential integration outputs. Third, multi-step integration is further analyzed by studying the impact of different candidate generation strategies in terms of tracking accuracy (Sect.\ref{ssec:sec4-3}).

To provide a generic evaluation while ensuring content diversity and representativity, video object tracking is performed over $10$ sequences (Tab.\ref{tab::sec4-0-tab1}) extracted from $4$ databases:  \texttt{bag}, \texttt{fish3} (denoted \texttt{fsh3}) and \texttt{octopus} (\texttt{octo}) from the Visual Object Tracking (VOT) database \cite{kristan2016vot}, \texttt{sleep1} (\texttt{sle1}) with albedo from MPI Sintel \cite{butler12eccv}, \texttt{lapa} from the laparoscopy dataset \cite{sznitman2012data} as well as \texttt{swan}, \texttt{bear}, \texttt{camel} (\texttt{caml}), \texttt{cows} and \texttt{flamingo} (\texttt{flam}) from the Densely Annotated VIdeo Segmentation (DAVIS) database \cite{perazzi2016benchmark}. As detailed in Tab.\ref{tab::sec4-0-tab1}, these sequences cover altogether many challenging situations such as complex non-rigid motion (\texttt{NR}), large displacement (\texttt{LD}), background clutter (\texttt{BC}), i.e. color similarities with background or between objects, dynamic background (\texttt{DB}) including moving background objects and camera viewpoint changes, scale variations (\texttt{SV}), partial occlusions (\texttt{PO}), thin structures (\texttt{TS}), illuminations changes and shadows (\texttt{IC}). Except for \texttt{lapa} whose ground-truth (GT) masks have been created from our own, all sequences were provided with associated GT masks indicating exact object delineations. 

\begin{table}
\scriptsize 
\centering \begin{tabular}{c|c|c|c|c|c|c|c|c|c|c|}
\cline{2-11}
 & img & obj & \texttt{NR} & \texttt{LD} & \texttt{BC} & \texttt{DB} & \texttt{SV} & \texttt{PO} & \texttt{TS} & \texttt{IC} \cr \hline
\multicolumn{1}{|c|}{\texttt{bag} \cite{kristan2016vot}} & 101 & 1 & x & x & & x & & & & x \cr \hline
\multicolumn{1}{|c|}{\texttt{fsh3} \cite{kristan2016vot}} & 101 & 1 & & x & & x & & & & x \cr \hline
\multicolumn{1}{|c|}{\texttt{octo} \cite{kristan2016vot}} & 51 & 1 & x & & x & x & x & & & \cr \hline
\multicolumn{1}{|c|}{\texttt{lapa} \cite{sznitman2012data}} & 81 & 1 & & & x & x & & & & x \cr \hline
\multicolumn{1}{|c|}{\texttt{sle1} \cite{butler12eccv}} & 50 & 3 & & & x & & x & & x & \cr \hline
\multicolumn{1}{|c|}{\texttt{swan} \cite{perazzi2016benchmark}} & 50 & 1 & & x & & x & & & & x \cr \hline
\multicolumn{1}{|c|}{\texttt{bear} \cite{perazzi2016benchmark}} & 82 & 1 & x & x & x & x & & x & & x \cr \hline
\multicolumn{1}{|c|}{\texttt{caml} \cite{perazzi2016benchmark}} & 90 & 1 & x & x & x & x & & x & x & x \cr \hline
\multicolumn{1}{|c|}{\texttt{cows} \cite{perazzi2016benchmark}} & 104 & 1 & x & x & & x & x & x & & x \cr \hline
\multicolumn{1}{|c|}{\texttt{flam} \cite{perazzi2016benchmark}} & 80 & 1 & x & x & x & x & & & x & x \cr \hline
\end{tabular}
\caption{Overview of sequences extracted from \cite{kristan2016vot,sznitman2012data,butler12eccv,perazzi2016benchmark} and used for object tracking experiments with associated sequence length, tracked object number and video attributes including complex non-rigid motion (\texttt{NR}), large displacement (\texttt{LD}), background clutter (\texttt{BC}), dynamic background (\texttt{DB}), scale variations (\texttt{SV}), partial occlusions (\texttt{PO}), thin structures (\texttt{TS}), illuminations changes and shadows (\texttt{IC}).} \vspace{-0.2cm}
\label{tab::sec4-0-tab1}
\end{table}

Video object tracking, also called semi-supervised video object segmentation task, consists in estimating for the whole sequence the exact location of a semantically meaningful free-shape region of interest (ROI) manually defined in one single image referred as reference frame. Once produced, tracking results are assessed for each pair $\{I_{ref},I_{n}\}$ with $n\hspace{-0.08cm}\neq \hspace{-0.08cm} ref$ based on three complementary measures. First, DICE scores measure the region-based segmentation similarity between estimated $X$ and GT $Y$ masks by computing $\frac{2|X\cap Y|}{|X|+|Y|}$. Then, contour-based precision $P_{c}$ and recall $R_{c}$ between estimated and GT masks can be estimated relying on bipartite graph matching to be robust to small inaccuracies \cite{martin2004learning}. In practice, we focus on the F-measure combining precision and recall using $\mathcal{F} = \frac{2P_{c}R_{c}}{P_{c}+R_{c}}$. Bi-partite matching is approximated using morphology operators, as in \cite{perazzi2016benchmark}. Finally, consistency-based assessment is performed relying on the percentage of pixels of $I_{ref}$ located inside the tracked ROI and whose belonging superpixel $\textbf{\textit{f}}_{ref}$ is consistent in terms of forward-backward binary consistency: \vspace{-0.8cm} \\

\begin{equation}
h_{n,ref}(h_{ref,n}(\textbf{\textit{f}}_{ref}))=\textbf{\textit{f}}_{ref}
\label{eq::sec4-0-eq1}
\end{equation} \vspace{-0.95cm} \\

In terms of computation time, performing RF-based matching followed by multi-step integration using steps $\{1,2,5,10,20\}$ on a sequence of 640$\times$360 frames such as \texttt{octo} with $500$ superpixels takes approximately 6 minutes per frame using a $3.1$GHz Intel Xeon CPU processor and Python implementation, without extensive code optimization. Processing time are reduced about $17\%$ when relying on $k$NN for unsupervised learning.

\subsection{Elementary superpixel matching}
\label{ssec:sec4-1}

Our first experiments consist in evaluating the proposed unsupervised learning-based superpixel matching  (Sect.\ref{sec:sec2}) between consecutive and distant frames against state-of-the-art methods. In this direction, ROI tracking is performed through direct integration (\texttt{DIR}) in the to-the-reference direction, i.e. relying on direct processing of image pairs $\{I_{n},I_{ref}\}$ without any sequential or multi-step combinations of pre-estimated superpixel matches. unsupervised learning-based matching using both $k$NN and RF classifiers is compared to three other methodologies: superpixel-to-superpixel matching using superpixel-wise average color (RGBm) and color histogram (RGBh) features, PatchMatch (PM) \cite{barnes2009patchmatch}, as well as optical flow through Farneb\"ack \cite{farneback2003two} and SIFT Flow \cite{liu2011sift}. Unsupervised learning-based matching works with $|\mathcal{F}|=|\mathcal{S}|=500$ superpixels per frame and employs $K_{a}\hspace{-0.09cm}=\hspace{-0.07cm}80$ context-rich multi-channel features computed with $\Phi\hspace{-0.07cm}=\hspace{-0.07cm}40$ as maximal radius and $w\hspace{-0.03cm}\in\hspace{-0.03cm}\{3,5,7\}$ as possible box sizes (Sect.\ref{ssec:sec2-4}). RF is made of $T\hspace{-0.08cm}=\hspace{-0.09cm}100$ trees whereas $k$NN relies on $5$ neighbors for queries. RGBm and RGBh use respectively average RGB colors and RGB histograms (using $10$ bins) as superpixel-wise features to give correspondences in $I_{ref}$ for each superpixel of $I_{n}$ in a nearest neighbor manner. As for unsupervised learning-based matching, RGBm and RGBh exploit images decomposed into $500$ superpixels. PM \cite{barnes2009patchmatch} is looking for the best patch matches using $9\hspace{-0.07cm}\times\hspace{-0.07cm}9$ windows with $6$ iterations including both propagation and random refinement steps. Farneb\"ack \cite{farneback2003two} and SIFT Flow \cite{liu2011sift} estimators are used using by-default parameters. 

Learning-based and superpixel-to-superpixel matching are performed once a groundtruth label is assigned to each superpixel of $I_{ref}$ to indicate its belonging to the ROI to be tracked. 50\% of the constituting pixels must be included into the ROI to label a superpixel as part of the object in $I_{ref}$. Label propagation can be then easily done at the superpixel level once to-the-reference superpixel pairings are obtained. Conversely, PM and optical flow estimators use dense to-the-reference fields to propagate labels at the pixel level from $I_{ref}$ to the whole sequence.

\begin{table}
\scriptsize
\hspace{-2.5cm}\begin{tabular}{c?c|c|c|c|c|c|c?c|c|c|c|c|c|c|}
\cline{2-15}
\multirow{3}{*}{} & \multicolumn{7}{c?}{DICE} & \multicolumn{7}{c|}{F-measure}  \cr \cline{2-15}
\multirow{3}{*}{} & \multicolumn{2}{c|}{spx matching\hspace{-0.8cm}} & \multirow{2}{*}{PM\cite{barnes2009patchmatch}} & \multicolumn{2}{c|}{optical flow} &  \multicolumn{2}{c?}{proposed \texttt{DIR}} &
\multicolumn{2}{c|}{spx matching} & \multirow{2}{*}{PM\cite{barnes2009patchmatch}} & \multicolumn{2}{c|}{optical flow} &  \multicolumn{2}{c|}{proposed \texttt{DIR}} \cr 
\cline{2-3} \cline{5-10} \cline{12-15}
& RGBm & RGBh & & Far\cite{farneback2003two} & SF\cite{liu2011sift} & $k$NN & RF & RGBm & RGBh & & Far\cite{farneback2003two} & SF\cite{liu2011sift} & $k$NN & RF \cr \hline
\multicolumn{1}{|l?}{\texttt{bag}} & 67.1 & 87.5 & 11.6 & 11.8 & 11.0 & \textbf{96.9} & \underline{92.8} & 59.3 & 82.6 & 10.7 & 10.1 & 10.6 & \textbf{97.8} & \underline{89.0} \cr \hline
\multicolumn{1}{|l?}{\texttt{fsh3}} & \textbf{89.8} & \underline{89.7} & 23.8 & 16.7 & 15.2 & 64.8 & 89.4 & \textbf{91.9} & \underline{91.4} & 27.4 & 21.4 & 18.7 & 67.3 & 81.5 \cr \hline
\multicolumn{1}{|l?}{\texttt{octo}} & 25.6 & 47.6 & \textbf{98.6} & \underline{96.5} & 96.4 & 84.8 & 85.5 & 17.2 & 32.1 & \textbf{95.2} & \underline{92.9} & \underline{92.9} & 75.6 & 74.9 \cr \hline
\multicolumn{1}{|l?}{\texttt{lapa}} & 65.8 & 67.7 & 70.4 & 61.1 & 48.5 & \textbf{89.2} & \underline{87.9} & 57.1 & 57.7 & 53.0 & 51.2 & 51.5 & \textbf{86.1} & \underline{85.3} \cr \hline
\multicolumn{1}{|l?}{\texttt{sl1.1}} & \underline{84.5} & 84.2 & 41.7 &35.2 & 32.5 & 82.9 & \textbf{94.9} & 81.2 & \underline{83.3} & 53.1 & 42.2 & 43.1 & 81.8 & \textbf{94.4} \cr \hline
\multicolumn{1}{|l?}{\texttt{sl1.2}} & 24.5 & 51.2 & 24.3 &13.4 & 11.8 &  \underline{82.0} & \textbf{89.4} & 21.6 & 44.6 & 50.4 & 34.1 & 24.4 &  \underline{82.4} & \textbf{90.3} \cr \hline
\multicolumn{1}{|l?}{\texttt{sl1.3}} & 38.5 & 51.2 & 13.5 & 7.95 & 7.75 &  \underline{66.7} &  \textbf{88.3} & 39.5 & 49.1 & 24.7 & 12.1 & 11.4 &  \underline{72.9} & \textbf{77.5} \cr \hline
\multicolumn{1}{|l?}{\texttt{swan}} &  \underline{91.0} & 89.2 & 78.5 & 84.9 & 83.9 & 90.5 & \textbf{91.3} &  \underline{87.4} & 83.9 & 62.6 & 70.7 & 69.2 & 85.7 & \textbf{88.0} \cr \hline
\multicolumn{1}{|l?}{\texttt{bear}} & 81.5 & 73.8 & 79.7 & \underline{85.6} & 84.0 & 84.3 & \textbf{89.1} & 57.7 & 46.5 & 56.3 & 64.5 & \underline{65.9} & 65.2 & \textbf{76.3} \cr \hline
\multicolumn{1}{|l?}{\texttt{camel}} & 58.5 & 49.3 & \textbf{85.2} & \textbf{85.2} & 80.5 & 67.5 & 70.3 & 45.6 & 38.2 & \textbf{72.4} & 64.3 & \underline{67.2} & 56.4 & 59.3 \cr \hline
\multicolumn{1}{|l?}{\texttt{cows}} & 64.2 & 67.5 & \textbf{89.6} & \textbf{89.6} & 84.4 & 16.8 & 87.1 & 32.2 & 36.9 & \textbf{77.3} & 68.3 & 60.4 & 15.5 & \underline{69.8} \cr \hline
\multicolumn{1}{|l?}{\texttt{flam}} & 48.2 & 52.4 & \textbf{77.5} & \underline{77.2} & 66.3 & 70.0 & 76.2 & 45.6 & 46.7 & 55.6 & 48.8 & 45.9 & \underline{56.3} & \textbf{67.2} \cr \hline \hline
\multicolumn{1}{|l?}{\textit{avg}} & \textit{61.6} & \textit{67.6} & \textit{57.9} & \textit{54.2} & \textit{51.9} & \textit{\underline{74.7}} & \textit{\textbf{86.9}} & \textit{53.0} & \textit{57.8} & \textit{53.2} & \textit{48.4} & \textit{46.8} & \textit{\underline{70.3}} & \textit{\textbf{80.3}} \cr \hline
\end{tabular}
\caption{DICE and F-measure scores for ROI tracking across $10$ sequences using direct integration (\texttt{DIR}), i.e. direct processing of image pairs $\{I_{ref},I_{n}\}$. Four methodologies are compared: superpixel-to-superpixel matching using superpixel-wise average color (RGBm) and color histogram (RGBh) features, PatchMatch (PM) \cite{barnes2009patchmatch}, optical flow through Farneb\"ack \cite{farneback2003two} and SIFT Flow \cite{liu2011sift} as well as the proposed unsupervised learning-based superpixel matching using $k$NN/RF classifiers. Bold and underline results indicate first and second best scores.} \vspace{-0.22cm} 
\label{tab::sec4-1-tab1}
\end{table}

DICE and F-measure scores temporally averaged across each of the previously described sequences are given Tab.\ref{tab::sec4-1-tab1} for each method. Bold and underline results indicate first and second best scores. Results indicate a good matching accuracy reached using the proposed unsupervised learning-based strategy for both consecutive and distant frames. On average, RF-based superpixel pairings provide the best direct tracking results with DICE and F-measure of $86.9$ and $80.3$, followed by $k$-NN-based results which reach $74.7$ and $70.3$. Both methods are significantly superior to the other state-of-the-art methods. Averaged DICE (F-measure) goes down to $67.6$ ($57.8$) and $61.6$ ($53.0$) for RGBh and RGBm respectively. Despite fairly good scores for \texttt{octo}, \texttt{caml}, \texttt{cows} and \texttt{flam}, PM and optical flow methods do not globally outperform unsupervised learning-based and superpixel-to-superpixel matching with averaged DICE (F-measure) of $57.9$ ($53.2$), $54.2$ ($48.4$) and $51.9$ ($46.8$) for PM, Farneb\"ack and SIFT Flow. 

\begin{figure}[t]
\centering \begin{tabular}{cccc}
\hspace{-0.4cm} \includegraphics[width=2.2cm]{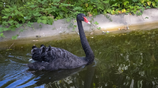} &
\hspace{-0.48cm} \includegraphics[width=2.2cm]{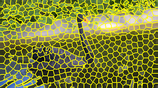} &
\hspace{-0.48cm} \includegraphics[width=2.2cm]{./figures/swan/034.png} &
\hspace{-0.48cm} \includegraphics[width=2.2cm]{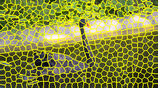} \vspace{-0.35cm} \cr
\hspace{-0.4cm} \scriptsize $I_{1}$ &
\hspace{-0.48cm} \scriptsize SLIC, $I_{1}$ &
\hspace{-0.48cm}\scriptsize $I_{34}$ &
\hspace{-0.48cm} \scriptsize SLIC, $I_{34}$ \cr
\hspace{-0.4cm} \includegraphics[width=2.2cm]{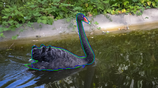} &
\hspace{-0.48cm} \includegraphics[width=2.2cm]{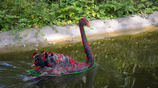} &
\hspace{-0.48cm} \includegraphics[width=2.2cm]{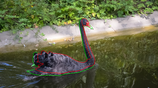} &
\hspace{-0.48cm} \includegraphics[width=2.2cm]{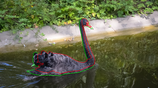} \vspace{-0.35cm} \cr
\hspace{-0.4cm} \scriptsize GT assignment &
\hspace{-0.48cm} \scriptsize PM \cite{barnes2009patchmatch} &
\hspace{-0.48cm} \scriptsize Far \cite{farneback2003two} &
\hspace{-0.48cm} \scriptsize SF \cite{liu2011sift} \cr
\hspace{-0.4cm} \includegraphics[width=2.2cm]{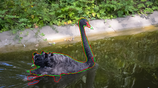} &
\hspace{-0.48cm} \includegraphics[width=2.2cm]{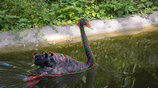} &
\hspace{-0.48cm} \includegraphics[width=2.2cm]{./figures/swan/dir-kNN-01-contours-034-th-50.png} &
\hspace{-0.48cm} \includegraphics[width=2.2cm]{./figures/swan/dir-RF-01-contours-034-th-50.png} \vspace{-0.35cm} \cr
\hspace{-0.4cm} \scriptsize RGBm &
\hspace{-0.48cm} \scriptsize RGBh &
\hspace{-0.48cm} \scriptsize $k$NN - \texttt{DIR} &
\hspace{-0.48cm} \scriptsize RF - \texttt{DIR} \cr
\hspace{-0.4cm} \includegraphics[width=2.2cm]{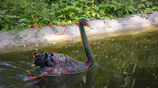} &
\hspace{-0.48cm} \includegraphics[width=2.2cm]{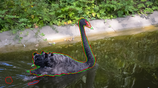} &
\hspace{-0.48cm} \includegraphics[width=2.2cm]{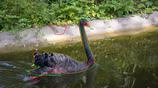} &
\hspace{-0.48cm} \includegraphics[width=2.2cm]{./figures/swan/msi-RF-01-contours-034-th-50.png} \vspace{-0.35cm} \cr
\hspace{-0.4cm} \scriptsize $k$NN - \texttt{SEQ} &
\hspace{-0.48cm} \scriptsize $k$NN - \texttt{MSI} &
\hspace{-0.48cm}\scriptsize RF - \texttt{SEQ} &
\hspace{-0.48cm} \scriptsize RF - \texttt{MSI} \cr
\hspace{-0.4cm} \includegraphics[width=2.2cm]{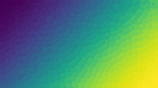} &
\hspace{-0.48cm} \includegraphics[width=2.2cm]{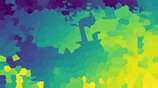} &
\hspace{-0.48cm} \includegraphics[width=2.2cm]{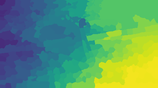} &
\hspace{-0.48cm} \includegraphics[width=2.2cm]{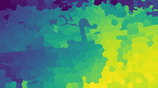} \vspace{-0.35cm} \cr
\hspace{-0.4cm} \scriptsize training &
\hspace{-0.48cm} \scriptsize RF - \texttt{DIR} pred. &
\hspace{-0.48cm} \scriptsize RF - \texttt{SEQ} pred. &
\hspace{-0.48cm} \scriptsize RF - \texttt{MSI} pred. \vspace{-0.1cm} \cr
\end{tabular}
\caption{ROI propagation for \texttt{swan} ($\{I_{1},I_{34}\}$) \cite{perazzi2016benchmark} with \texttt{DIR}, \texttt{SEQ} and \texttt{MSI} (steps $\{1,2,5,10,20,30\}$) integrations using $k$NN and RF. Results are compared with: superpixel-to-superpixel matching with average color (RGBm) and color histogram (RGBh) features, PatchMatch (PM) \cite{barnes2009patchmatch}, optical flow through Farneb\"ack (Far) \cite{farneback2003two} and SIFT Flow (SF) \cite{liu2011sift}. Blue boundaries in $I_{1}$ indicate superpixel labelling resulting from GT assignment. Green and red boundaries correspond to groundtruth (GT) and estimated tracking results. The last raw displays training ($I_{1}$) and prediction ($I_{34}$) masks resulting from \texttt{DIR}, \texttt{SEQ} and \texttt{MSI} integrations of RF-based elementary pairings.}
\label{fig::sec4-1-fig1}
\end{figure}

These findings are illustrated visually Fig.\ref{fig::sec4-1-fig1} for the pair $\{I_{1},I_{34}\}$ of \texttt{swan}. Red and green boundaries denote propagated and GT ROI location. We can notice that PM, Farneb\"ack and SIFT Flow under-estimate the area covered by the swan, especially for the neck and near the water. RGBm, RGBh as well as RF and $k$NN-based direct (\texttt{DIR}) superpixel pairings provide clearly better contours despite the tendency to propagate the ROI outside the swan area due to shadows and color similarities with background. One drawback with superpixel-based methods is the lack of boundary adherence which may not suit perfectly the object to be tracked. This aspect is revealed for $I_{1}$ by the blue boundaries which indicate the superpixel labelling resulting from GT assignment. However, Tab.\ref{tab::sec4-1-tab1} demonstrates that this limitation is compensated by robust superpixel matching heuristics compared to straithforward pixel-wise matching and mask propagation which rely on a regular decomposition of the image grid without enough context considerations. Finally, Fig.\ref{fig::sec4-1-fig1} shows a more accurate propagation achieved with multi-step integration of learning-based superpixel pairings, especially with RF. It tends to indicate the ability of unsupervised learning-based superpixel matching to provide a reliable and accurate enough elementary estimator towards efficient long-term multi-step matching and tracking. The performance achieved with multi-step integration is more deeply demonstrated in the next section.

\subsection{Long-term superpixel tracking}
\label{ssec:sec4-2}

Long-term ROI tracking resulting from direct (\texttt{DIR}), sequential (\texttt{SEQ}) and multi-step (\texttt{MSI}) integrations are compared based on unsupervised learning-based superpixel matching whose accuracy against state-of-the-art methods has been demonstrated in Sect.\ref{ssec:sec4-1} with $k$NN and RF classifiers. \texttt{MSI} is applied with $L\hspace{-0.07cm}=\hspace{-0.07cm}200$ maximal step sequences per image pairs. Only step sequences whose length is less than or equal to $K_{max}\hspace{-0.07cm}=\hspace{-0.07cm}7$ are kept to prevent from motion drift (Sect.\ref{ssec:sec3-2}). $k$NN and RF-based elementary multi-step superpixel matches are obtained with steps $\{1,2,5,10,20\}$ for \texttt{octo}, \texttt{swan} and \texttt{sle1} and $\{1,2,5,10,20,30,50\}$ for longer sequences (\texttt{fsh3}, \texttt{bag}, \texttt{lapa}, \texttt{bear}, \texttt{caml}, \texttt{cows} and \texttt{flam}). Context-rich features are estimated using the same parameters as in Sect.\ref{ssec:sec4-1}. Majority voting (Eq.\ref{eq::sec3-3-eq1}) focuses only on superpixel candidates generated in both to/from-the-reference directions to improve forward-backward consistency (see Sect.\ref{ssec:sec4-3} for further details).

\begin{table}
\scriptsize
\hspace{-3.7cm} \begin{tabular}{c?c|c|c?c|c|c?c|c|c?c|c|c?c|c|c?c|c|c|}
\cline{2-19}
\multirow{3}{*}{} & \multicolumn{6}{c?}{DICE} & \multicolumn{6}{c?}{F-measure}  & \multicolumn{6}{c|}{consistency} \cr \cline{2-19}
& \multicolumn{3}{c?}{$k$NN} & \multicolumn{3}{c?}{RF} & \multicolumn{3}{c?}{$k$NN} & \multicolumn{3}{c?}{RF} & \multicolumn{3}{c?}{$k$NN} & \multicolumn{3}{c|}{RF} \cr \cline{2-19}
& \texttt{DIR} & \texttt{SEQ} & \texttt{MSI} & \texttt{DIR} & \texttt{SEQ} & \texttt{MSI} & \texttt{DIR} & \texttt{SEQ} & \texttt{MSI} & \texttt{DIR} & \texttt{SEQ} & \texttt{MSI} & \texttt{DIR} & \texttt{SEQ} & \texttt{MSI} & \texttt{DIR} & \texttt{SEQ} & \texttt{MSI} \cr \hline
\multicolumn{1}{|l?}{\texttt{bag}} & 
96.9 & \textbf{\underline{97.7}} & \textbf{\underline{97.7}} & 
92.8 & 74.9 & \textbf{92.9} & 
97.8 & \textbf{\underline{99.7}} & 99.4 &
\textbf{89.0} & 65.9 & 88.3 &
\textbf{\underline{39.2}} & 12.1 & 32.1 &
\textbf{30.2} & 12.2 & 27.8 \cr \hline
\multicolumn{1}{|l?}{\texttt{fsh3}} & 
64.8 & \textbf{90.5} & 85.7 & 
89.4 & 91.1 & \textbf{\underline{92.0}} &
67.3 & \textbf{92.6} & 81.5 &
91.8 & 93.9 & \textbf{\underline{95.4}} &
56,5 & 37,6 & \textbf{\underline{74,3}} &
\textbf{69.5} & 38.8 & 58.7 \cr \hline
\multicolumn{1}{|l?}{\texttt{octo}} & 84.8 & 91.3 & \textbf{\underline{93.0}} & 
85.5 & 91.3 & \textbf{92.8} &
75.6 & 83.6 & \textbf{\underline{86.3}} &
74.9 & 83.6 & \textbf{86.0} &
\textbf{\underline{84,6}} & 67,5 & 76,6 &
\textbf{82.9} & 67,1 & 75,7 \cr \hline
\multicolumn{1}{|l?}{\texttt{lapa}} & 
89.2 & 86.5 & \textbf{\underline{92.8}} & 
87.9 & 88.3 & \textbf{\underline{92.8}} &
86.1 & 86.2 & \textbf{\underline{95.0}} &
85.3 & 86.8 & \textbf{94.8} &
96.1 & 68.9 & \textbf{\underline{96.8}} &
90.2 & 64.6 & \textbf{91.5} \cr \hline
\multicolumn{1}{|l?}{\texttt{sl1.1}} & 
82.9 & 95.3 & \textbf{95.6} & 
94.9 & 95.5 & \textbf{\underline{95.9}} &
81.8 & 93.2 & \textbf{94.8} &
94.4 & 94.0 & \textbf{\underline{95.5}} &
89,2 & 63,4 & \textbf{\underline{96,8}} &
\textbf{94,5} & 76,1 & 93,4 \cr \hline
\multicolumn{1}{|l?}{\texttt{sl1.2}} & 
82.0 & 80.9 & \textbf{90.3} & 
89.4 & 80.5 & \textbf{\underline{90.9}} &
82.4 & 77.9 & \textbf{93.0} &
90.3 & 73.1 & \textbf{\underline{93.9}} &
76.0 & 62,4 & \textbf{\underline{91,6}} &
84,9 & 61,6 & \textbf{91,2} \cr \hline
\multicolumn{1}{|l?}{\texttt{sl1.3}} & 
66.7 & \textbf{92.4} & 92.2 & 
88.3 & 92.4 & \textbf{\underline{94.2}} &
72.9 & 76.7 & \textbf{78.1} &
77.5 & 76.7 & \textbf{\underline{81.6}} &
86.7 & 78.1 & \textbf{95.0} &
87.7 & 75.6 & \textbf{\underline{100}} \cr \hline
\multicolumn{1}{|l?}{\texttt{swan}} & 
90.5 & 85.0 & \textbf{\underline{93.5}} &
91.3 & 86.9 & \textbf{93.4} &
85.7 & 73.9 & \textbf{93.6} &
88.0 & 78.4 & \textbf{\underline{93.8}} &
\textbf{\underline{86.1}} & 57.7 & 83.0 &
\textbf{82.3} & 59.3 & 77.7 \cr \hline
\multicolumn{1}{|l?}{\texttt{bear}} & 
84.3 & \textbf{87.8} & 87.7 &
89.1 & 87.0 & \textbf{\underline{92.5}} &
65.2 & 68.2 & \textbf{72.2} &
76.3 & 68.9 & \textbf{\underline{82.1}} &
65.6 & 61.2 & \textbf{\underline{68.5}} &
62.7 & 62.1 & \textbf{67.2} \cr \hline
\multicolumn{1}{|l?}{\texttt{camel}} & 
67.5 & 76.6 & \textbf{76.9} &
70.3 & 77.7 & \textbf{\underline{79.6}} &
56.4 & 64.7 & \textbf{67.2} &
59.3 & 63.3 & \textbf{\underline{69.1}} &
\textbf{58.7} & 41.5 & 51.8 &
\textbf{\underline{63.8}} & 42.3 & 54.7 \cr \hline
\multicolumn{1}{|l?}{\texttt{cows}} & 
16.8 & 80.1 & \textbf{87.3} &
87.1 & 79.1 & \textbf{\underline{89.1}} &
15.5 & 59.6 & \textbf{66.8} &
69.8 & 61.4 & \textbf{\underline{74.1}} &
15.5 & 39.6 & \textbf{\underline{68.5}} &
75.8 & 39.3 & \textbf{67.6} \cr \hline
\multicolumn{1}{|l?}{\texttt{flam}} & 
70.0 & 63.5 & 76.7 &
76.2 & 66.5 & \textbf{\underline{80.8}} &
56.3 & 62.5 & 67.1 &
67.2 & 62.8 & \textbf{\underline{71.9}} &
55.4 & 34.9 & 63.4 &
43.0 & 39.7 & \textbf{\underline{55.0}} \cr \hline \hline
\multicolumn{1}{|l?}{\textit{avg}} & 
\textit{74.7} & \textit{85.6} & \textit{\textbf{89.1}} &
\textit{86.9} & \textit{84.3} & \textit{\textbf{\underline{90.6}}} &
\textit{70.3} & \textit{78.2} & \textit{\textbf{82.9}} &
\textit{80.3} & \textit{75.7} & \textit{\textbf{\underline{85.5}}} &
\textit{67.5} & \textit{52.1} & \textit{\textbf{\underline{74.9}}} &
\textit{\textbf{72.3}} & \textit{53.2} & \textit{71.7}\cr \hline
\end{tabular}
\caption{DICE, F-measure and consistency scores for ROI tracking across $10$ sequences. We compare direct (\texttt{DIR}), sequential (\texttt{SEQ}) and multi-step (\texttt{MSI}) integration based on unsupervised learning-based superpixel matching using $k$NN and RF. Bold results indicate the best performance between \texttt{DIR}, \texttt{SEQ} and \texttt{MSI}. Underline scores highlight best results between $k$NN and RF-based methods.}
\label{tab::sec4-2-tab1}
\end{table}

Tab.\ref{tab::sec4-2-tab1} presents temporally averaged metrics (DICE, F-measure and consistency scores) obtained by \texttt{DIR}, \texttt{SEQ} and \texttt{MSI} across all sequences using $k$NN and RF. Except for consistency scores when relying on RF, Tab.\ref{tab::sec4-2-tab1} confirms that \texttt{MSI} is the best integration strategy towards long-term superpixel tracking compared to \texttt{DIR} and \texttt{SEQ}. For instance, RF and $k$NN-based \texttt{MSI} reach the highest DICE scores with $90.6$ and $89.1$ in comparison to $84.3$ ($86.9$) and $85.6$ ($74.7$) obtained with RF and $k$NN-based \texttt{SEQ} (\texttt{DIR}). Second and third positions in terms of DICE and F-measure vary depending on the classifier. \texttt{SEQ} outperforms \texttt{DIR} for $k$NN whereas RF exhibits the opposite behavior. Except for \texttt{MSI} in terms of consistency and \texttt{SEQ} for F-measure, another main finding is that RF-based elementary matches usually make better long-term tracking than $k$NN-based pairings, as one expects.

\begin{figure*}
\hspace{-2.5cm} \begin{tabular}{ccccc}
\hspace{-0.5cm} &
\hspace{-0.3cm} \small $k$NN &
\hspace{-0.7cm} \small RF &
\hspace{-0.7cm} \small $k$NN &
\hspace{-0.7cm} \small RF \vspace{-0.18cm} \cr
\hspace{-0.5cm} \small \rotatebox{90}{\textcolor{white}{-------} \texttt{lapa} \cite{sznitman2012data}} &
\hspace{-0.3cm} \includegraphics[width=4.3cm]{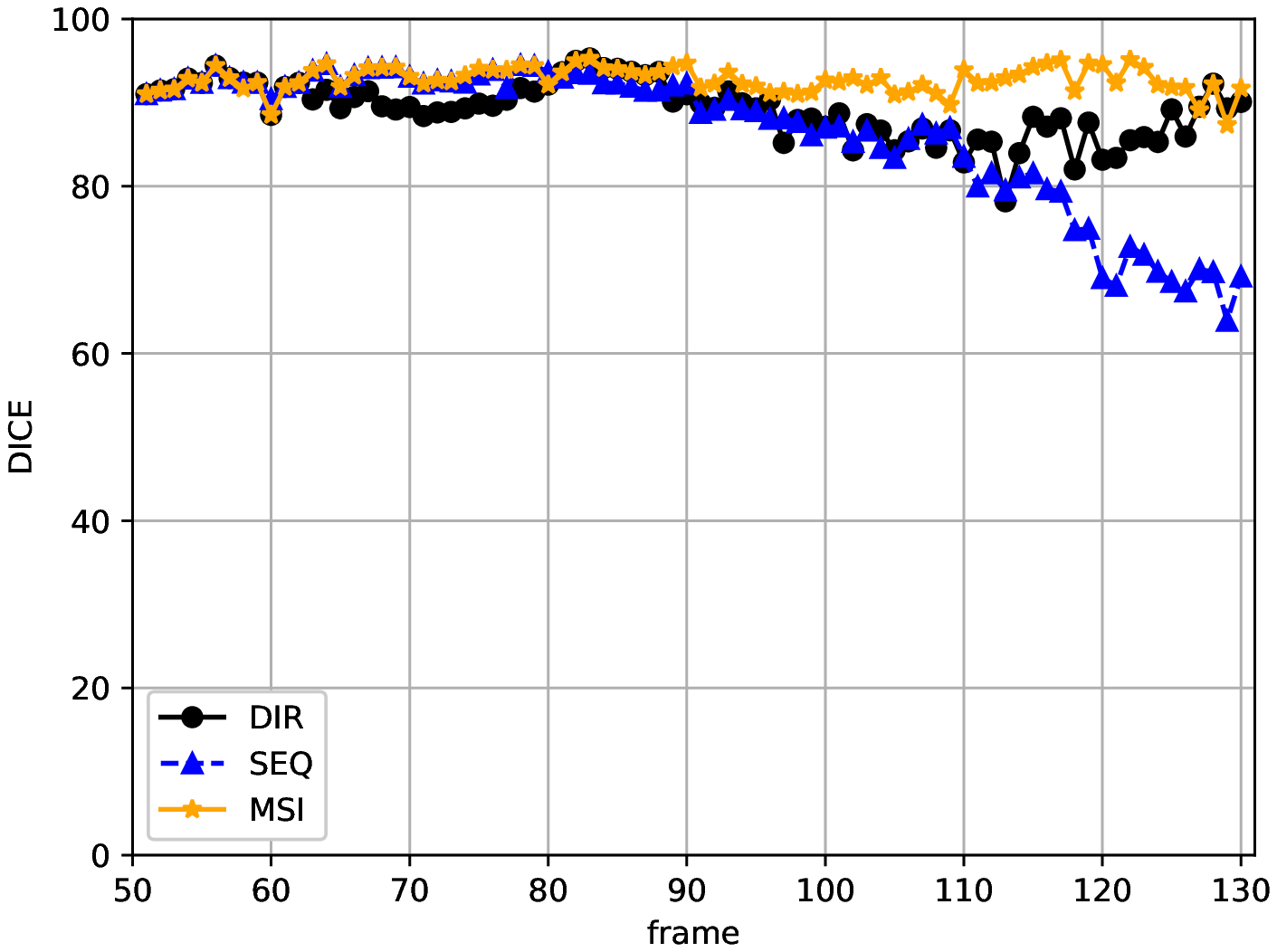} &
\hspace{-0.7cm} \includegraphics[width=4.3cm]{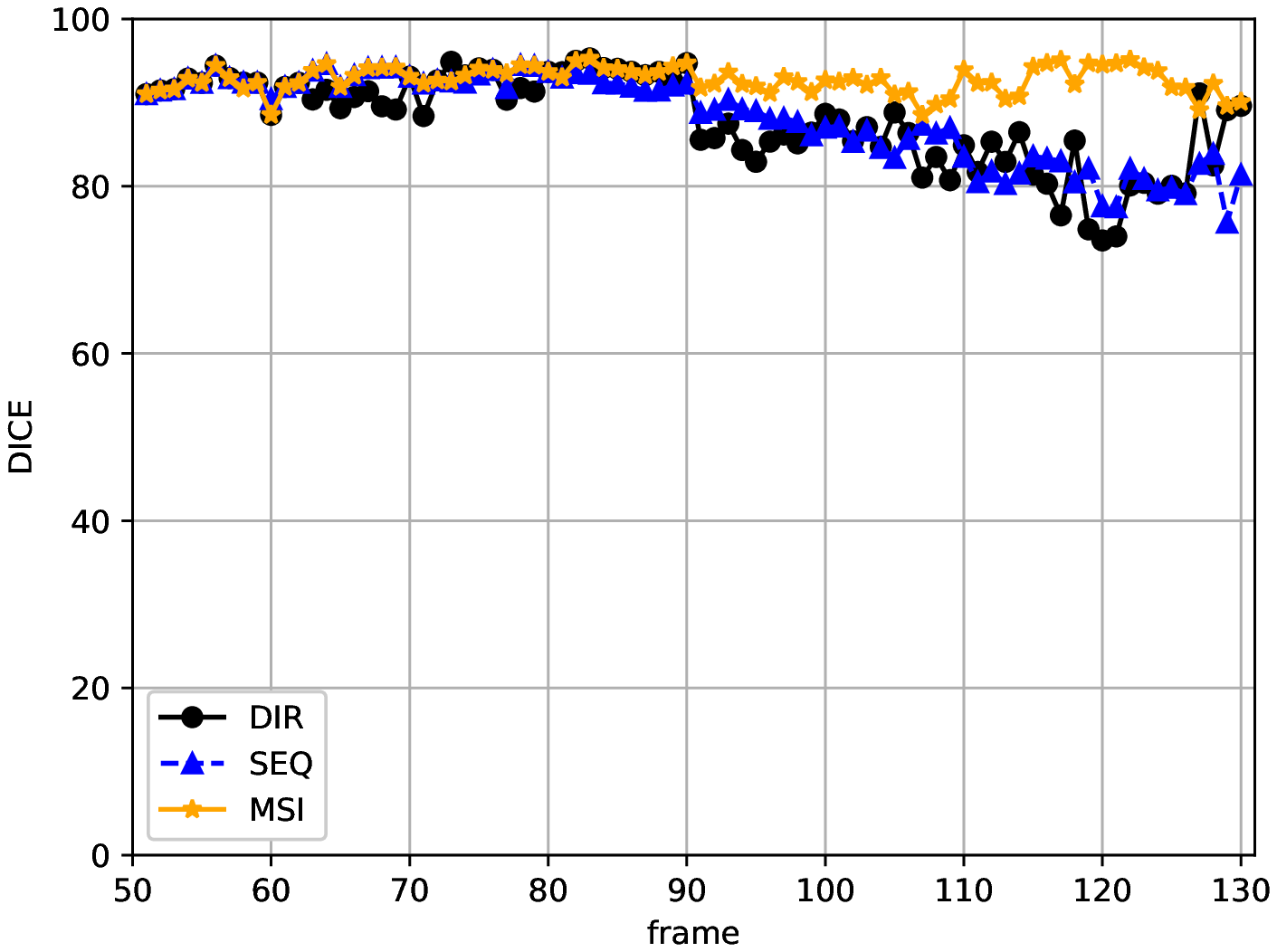} &
\hspace{-0.7cm} \includegraphics[width=4.3cm]{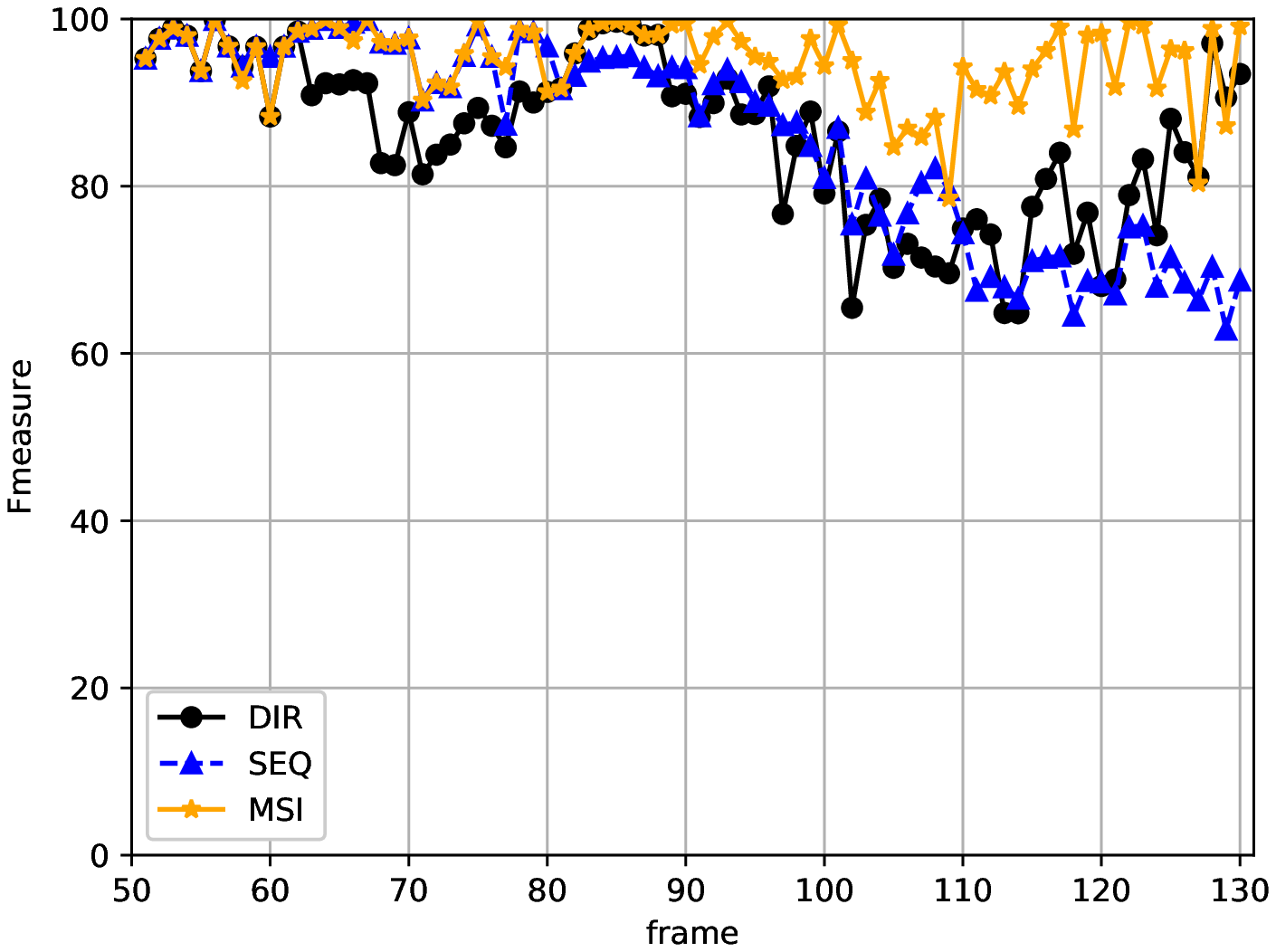} &
\hspace{-0.7cm} \includegraphics[width=4.3cm]{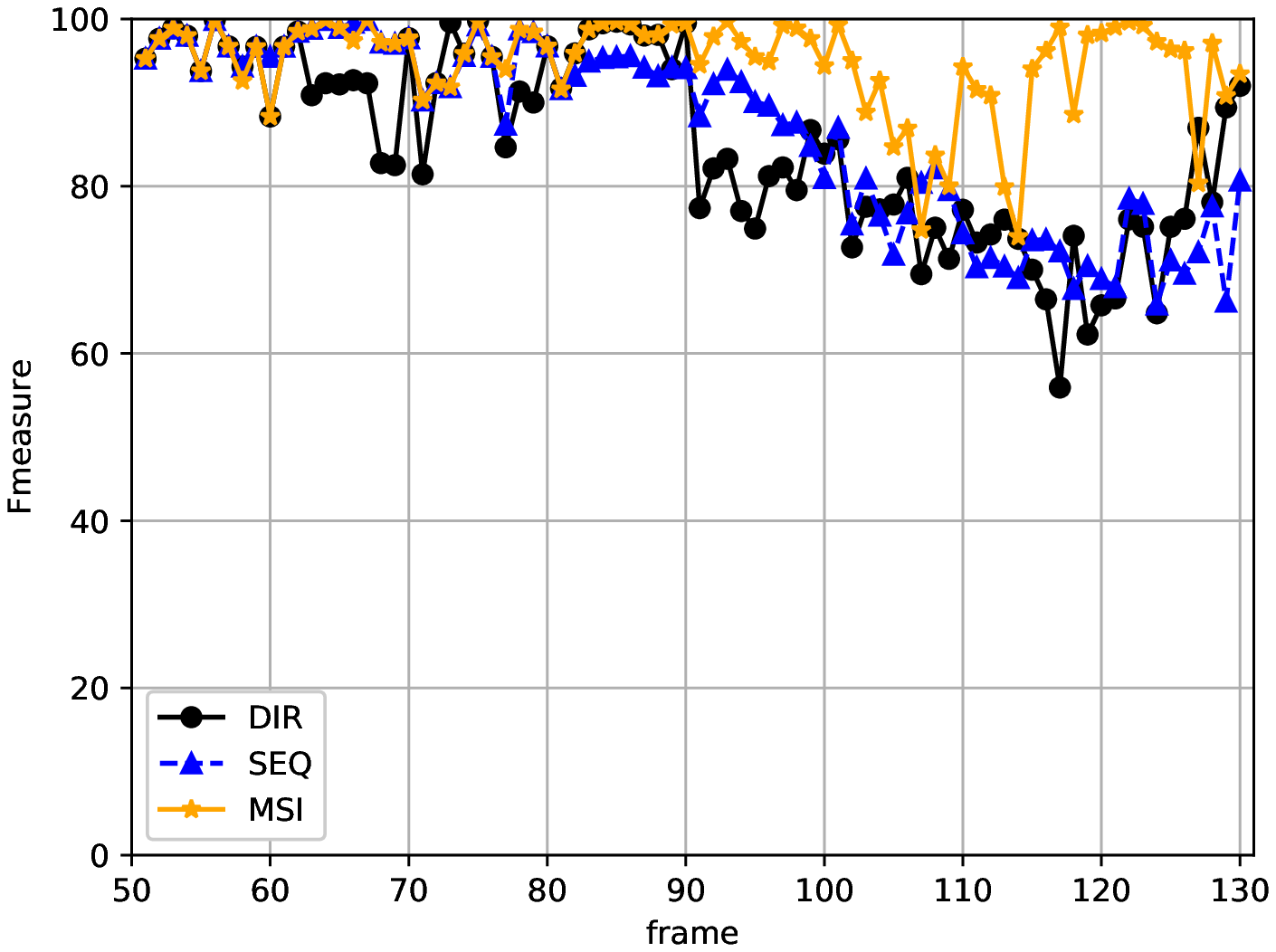} \cr
\hspace{-0.5cm} \small \rotatebox{90}{\textcolor{white}{------} \texttt{sle1.2} \cite{butler12eccv}} &
\hspace{-0.3cm} \includegraphics[width=4.3cm]{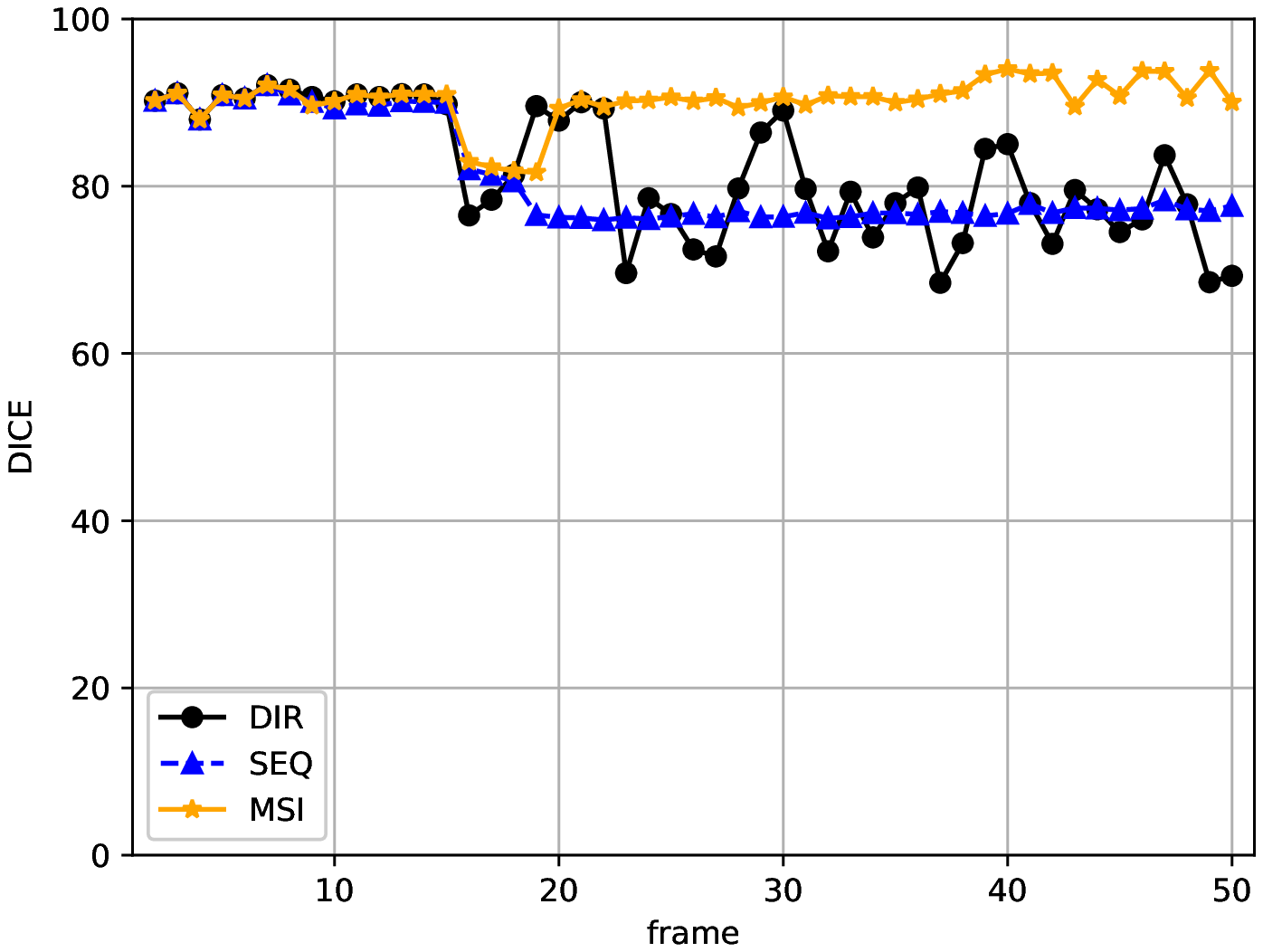} &
\hspace{-0.7cm} \includegraphics[width=4.3cm]{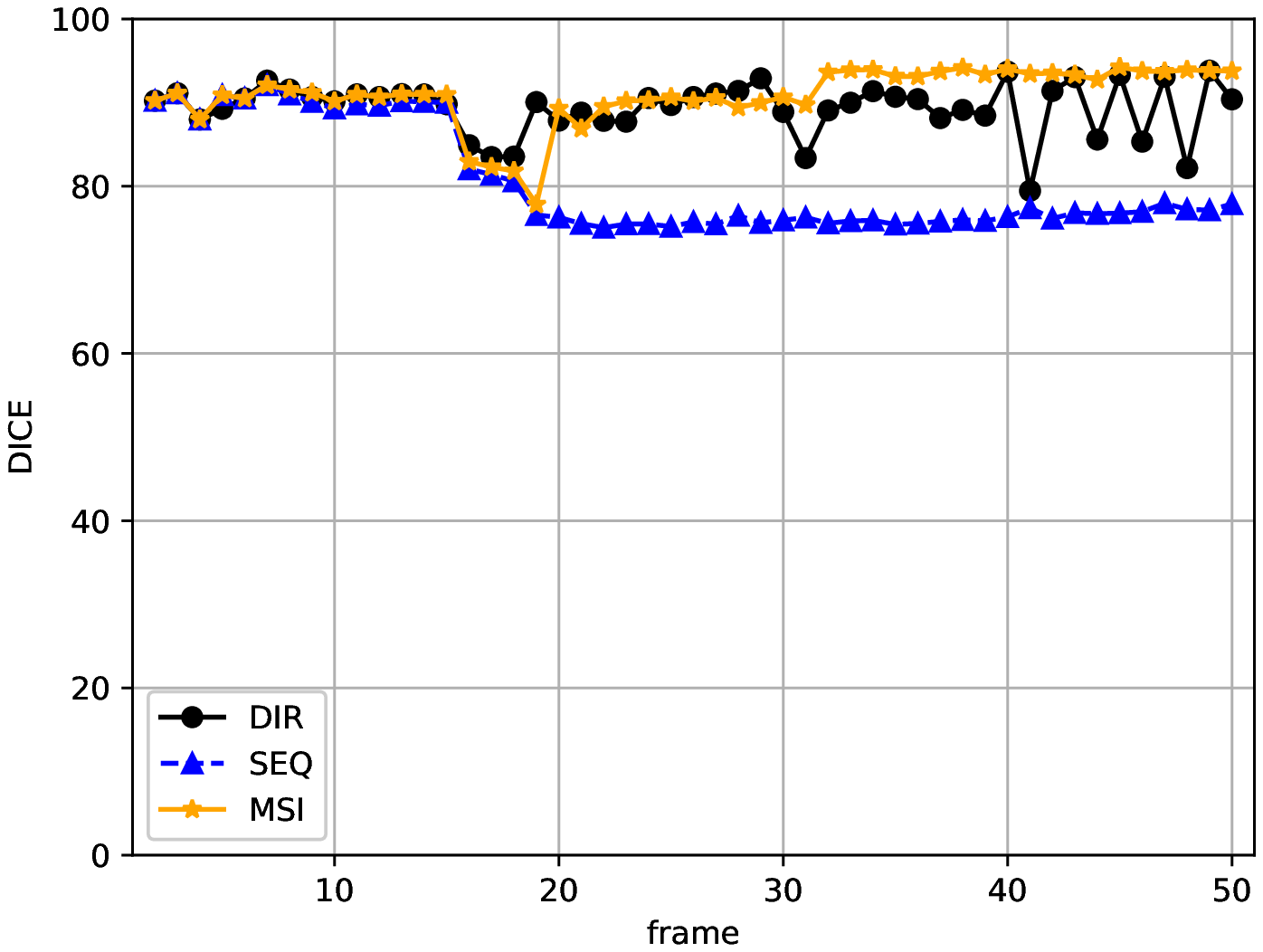} &
\hspace{-0.7cm} \includegraphics[width=4.3cm]{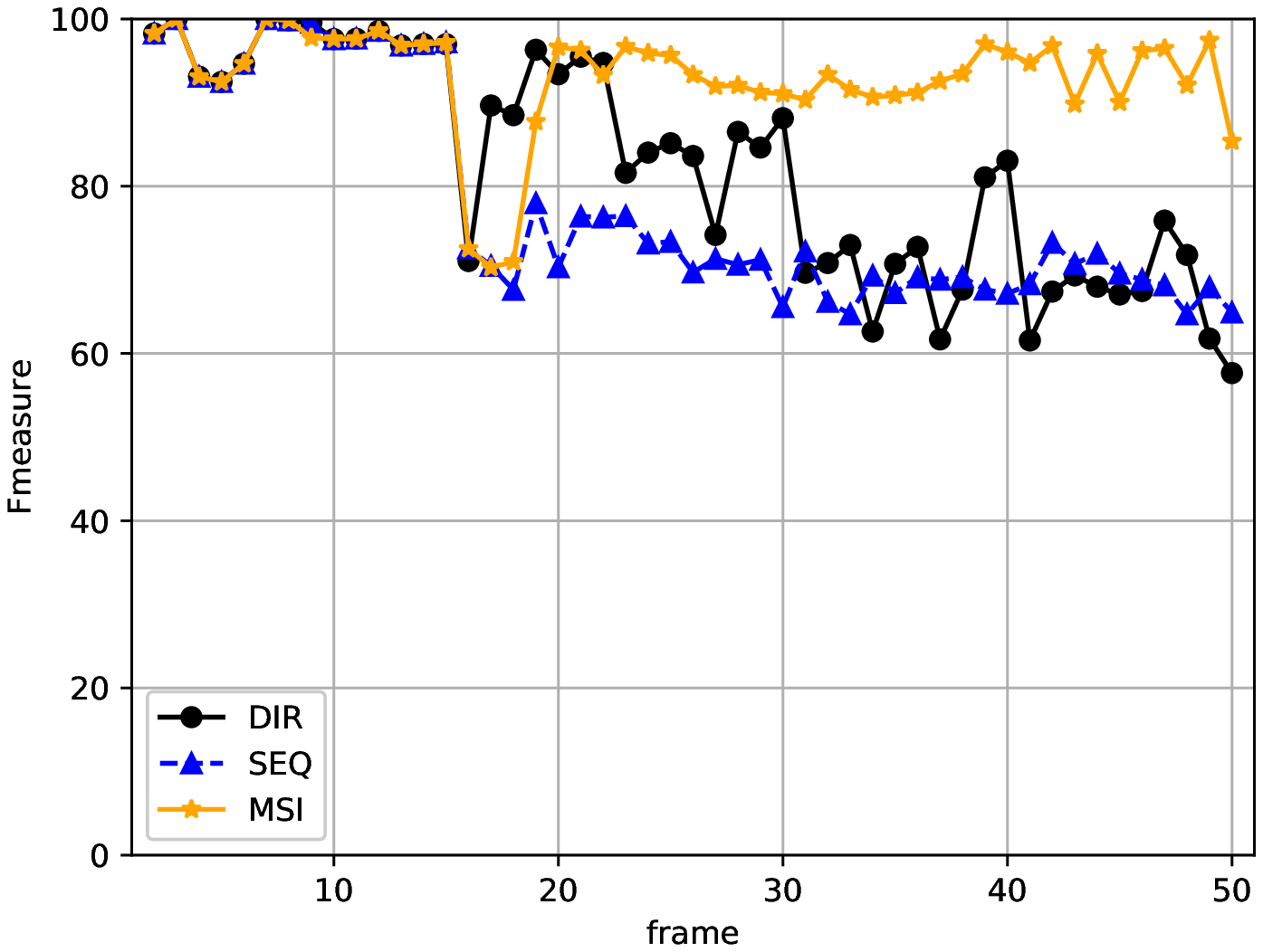} &
\hspace{-0.7cm} \includegraphics[width=4.3cm]{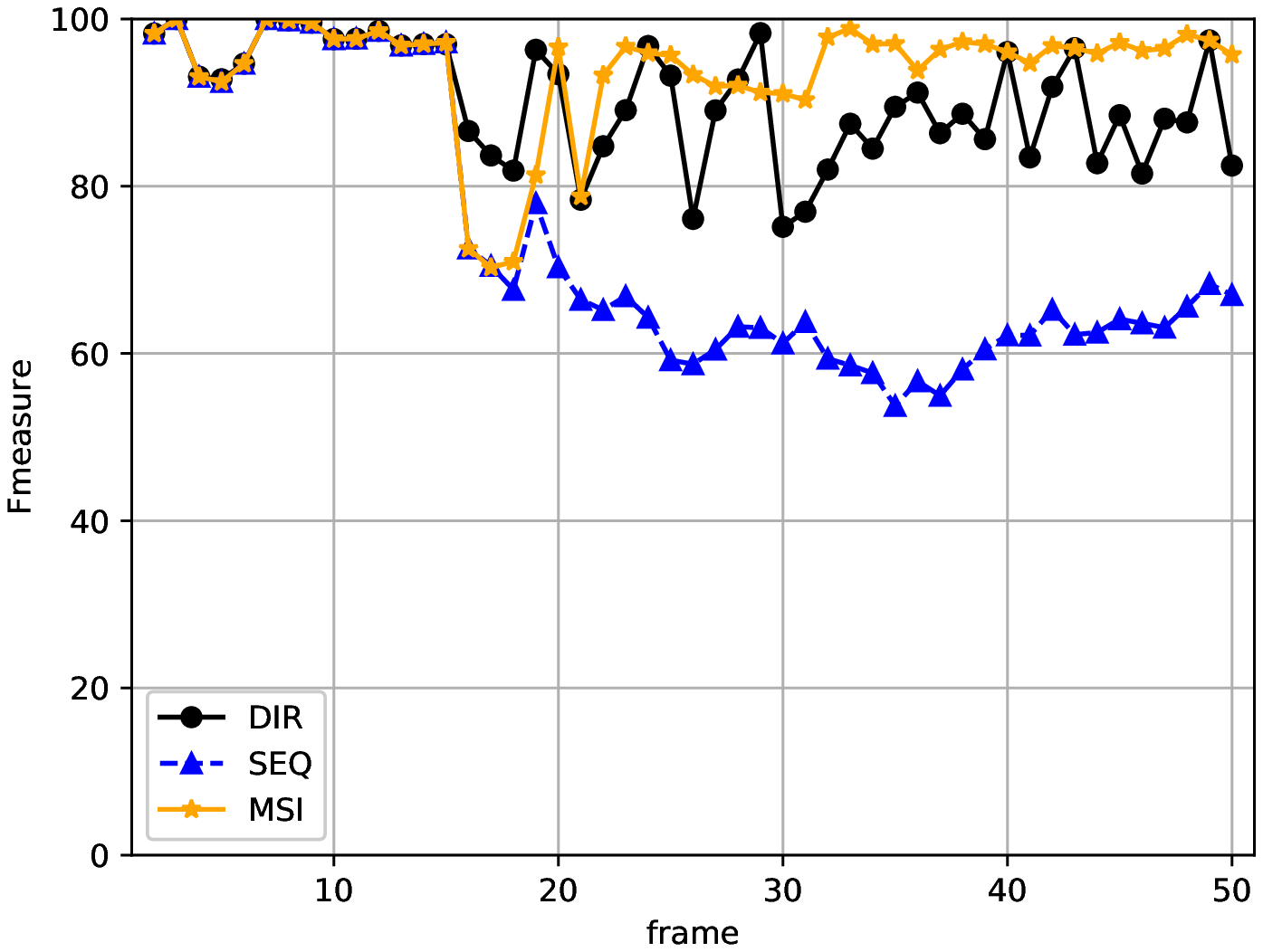} \cr
\hspace{-0.5cm} \small \rotatebox{90}{\textcolor{white}{--------} \texttt{octo} \cite{kristan2016vot}} &
\hspace{-0.3cm} \includegraphics[width=4.3cm]{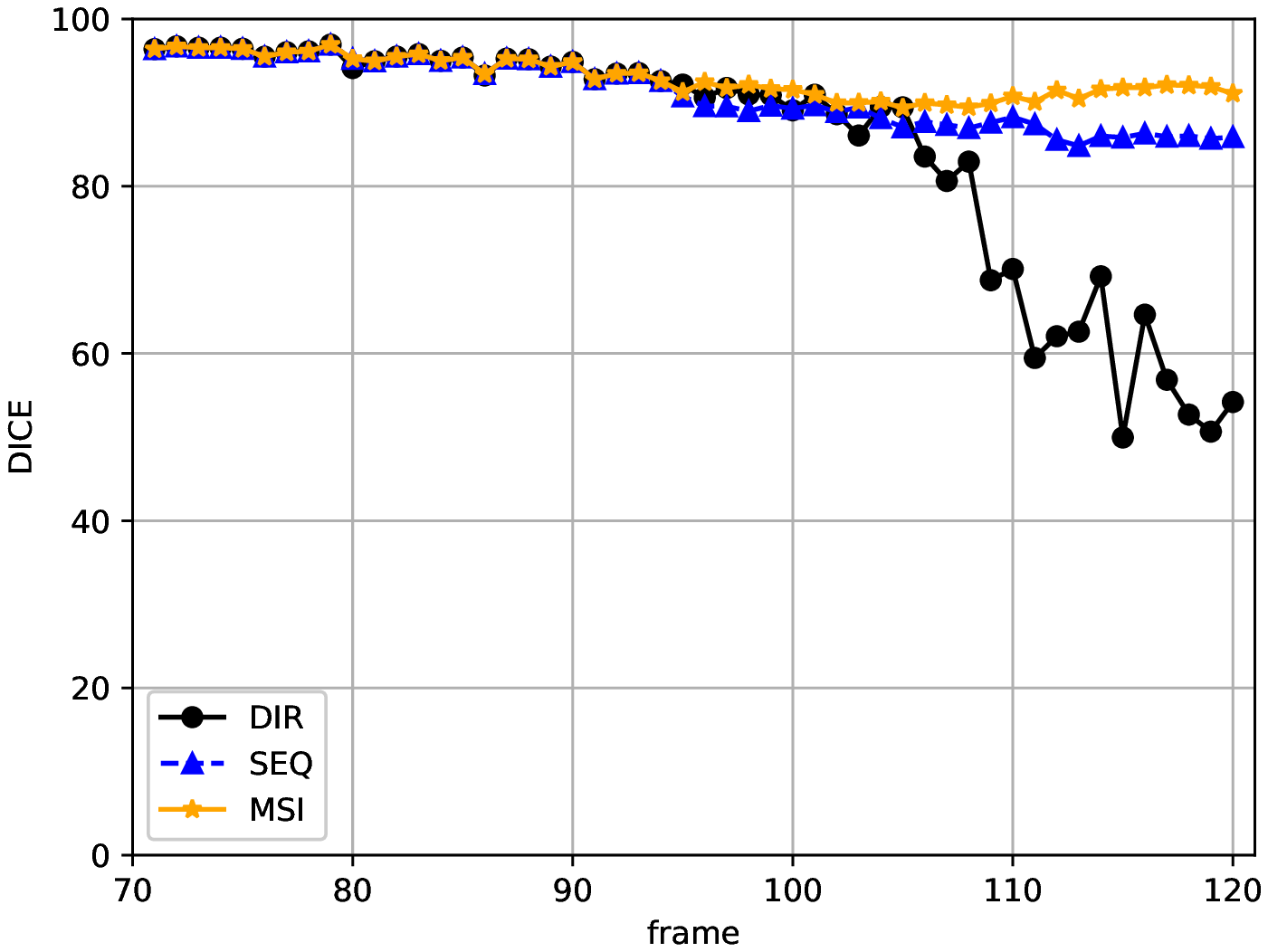} &
\hspace{-0.7cm} \includegraphics[width=4.3cm]{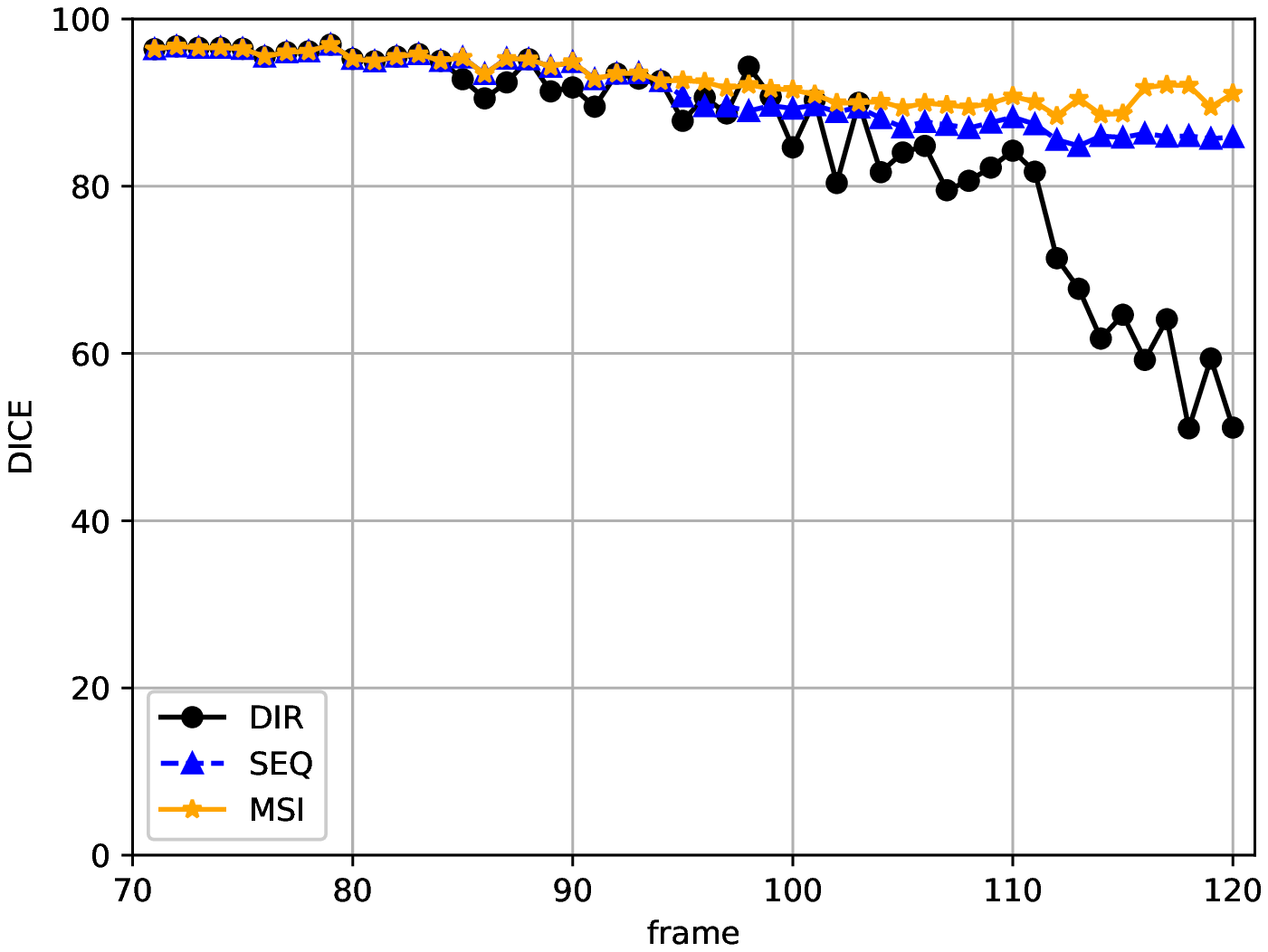} &
\hspace{-0.7cm} \includegraphics[width=4.3cm]{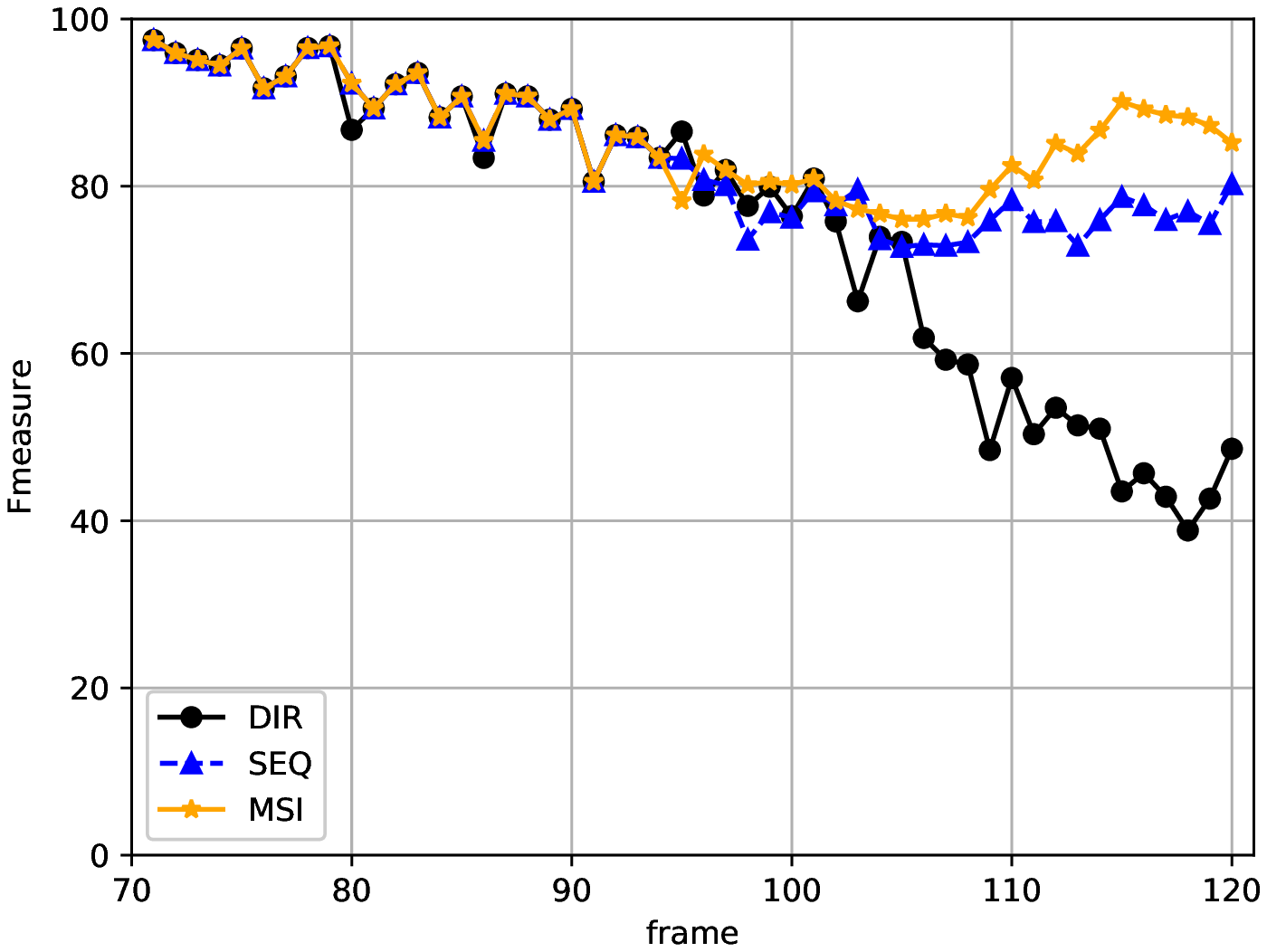} &
\hspace{-0.7cm} \includegraphics[width=4.3cm]{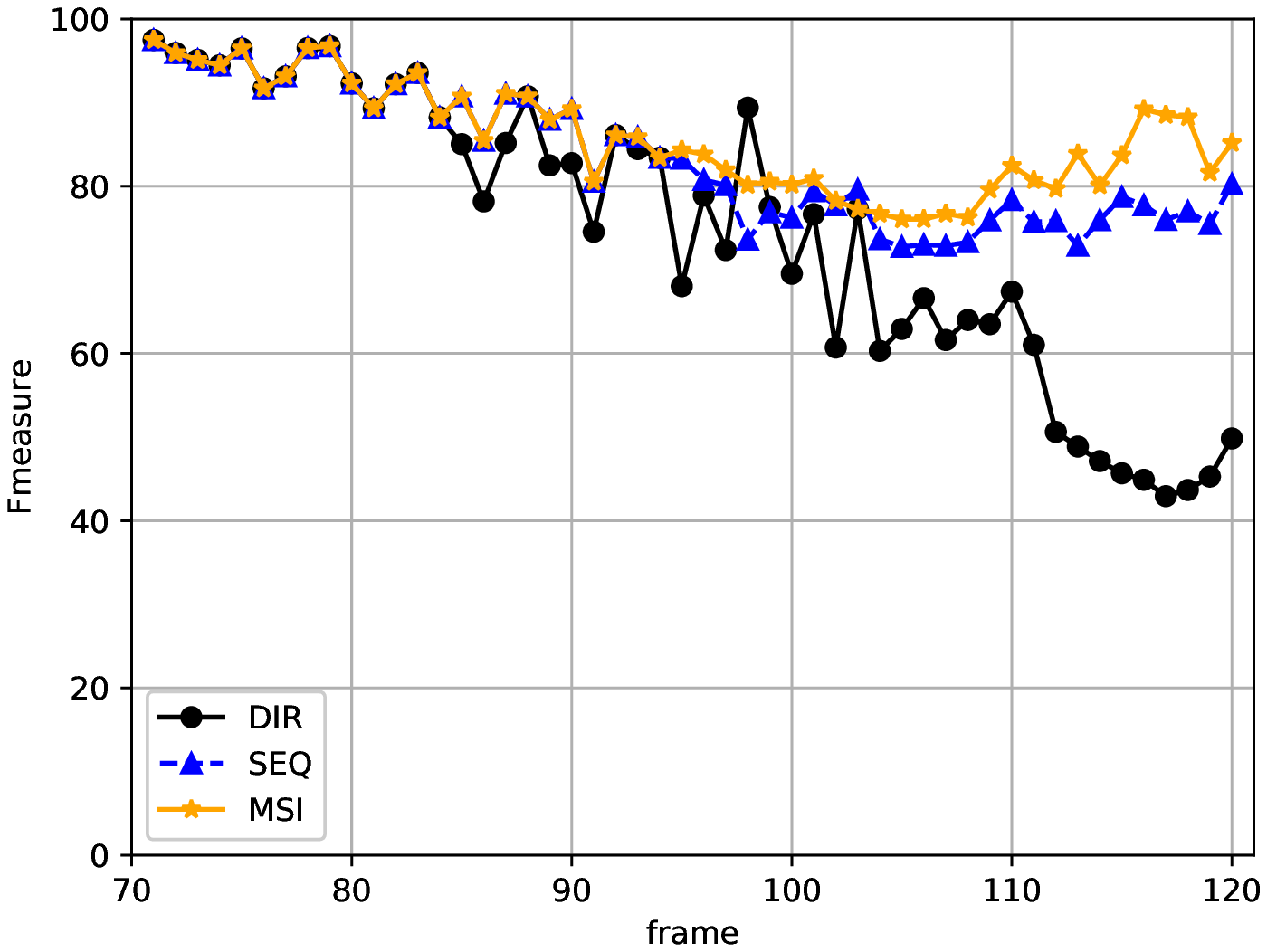} \cr
\hspace{-0.5cm} \small \rotatebox{90}{\textcolor{white}{---------} \texttt{swan} \cite{perazzi2016benchmark}} &
\hspace{-0.3cm} \includegraphics[width=4.3cm]{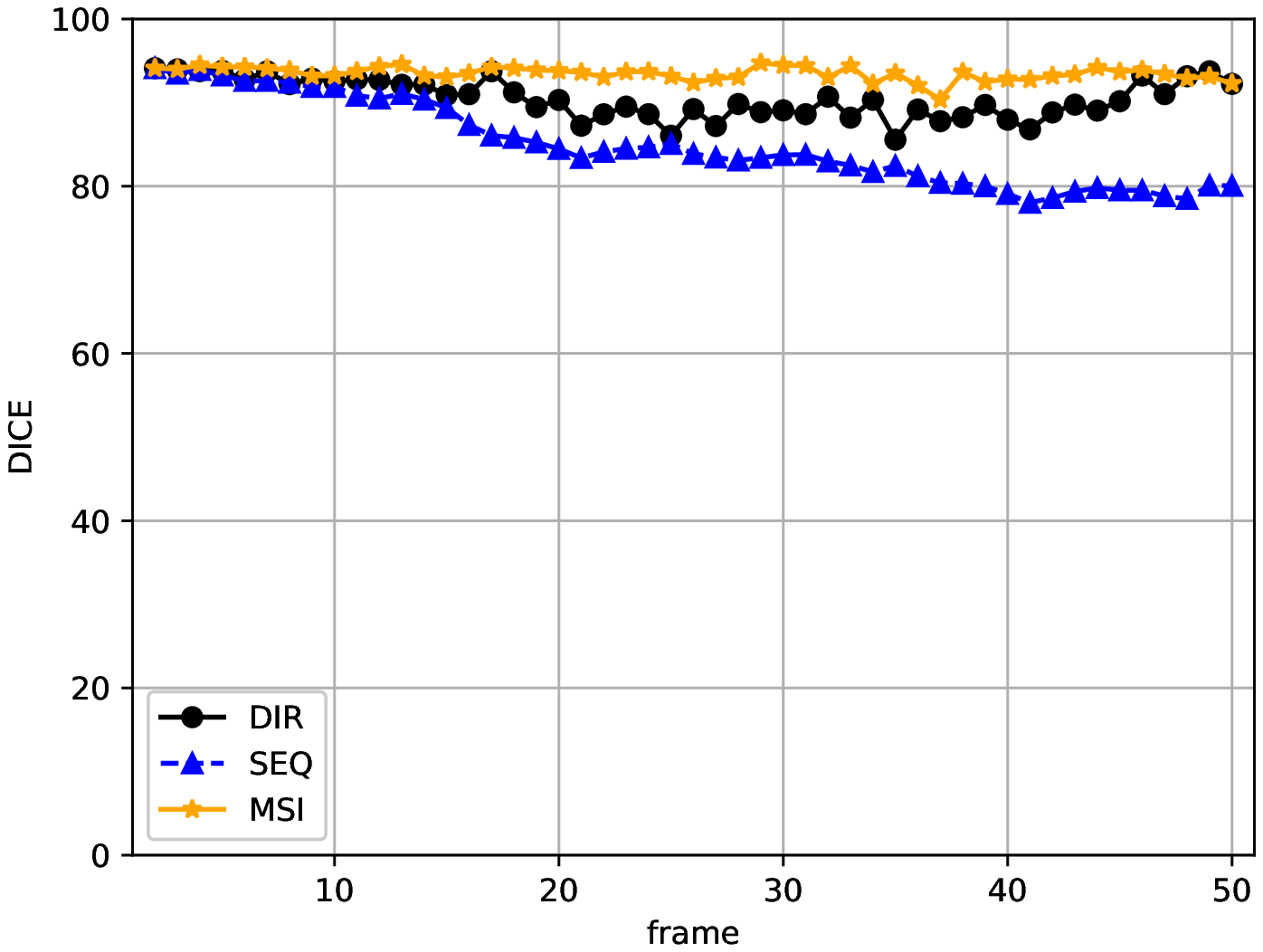} &
\hspace{-0.7cm} \includegraphics[width=4.3cm]{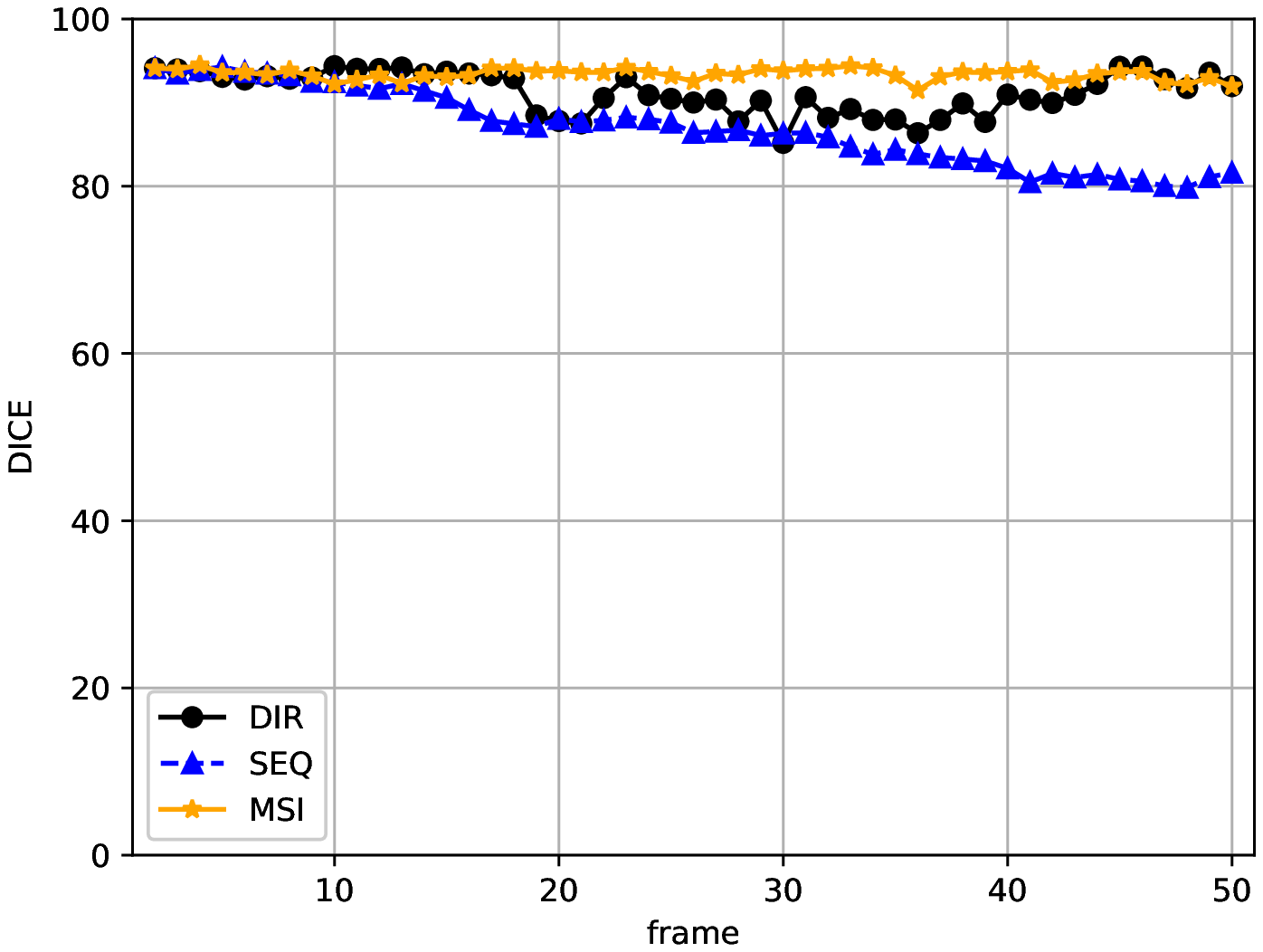} &
\hspace{-0.7cm} \includegraphics[width=4.3cm]{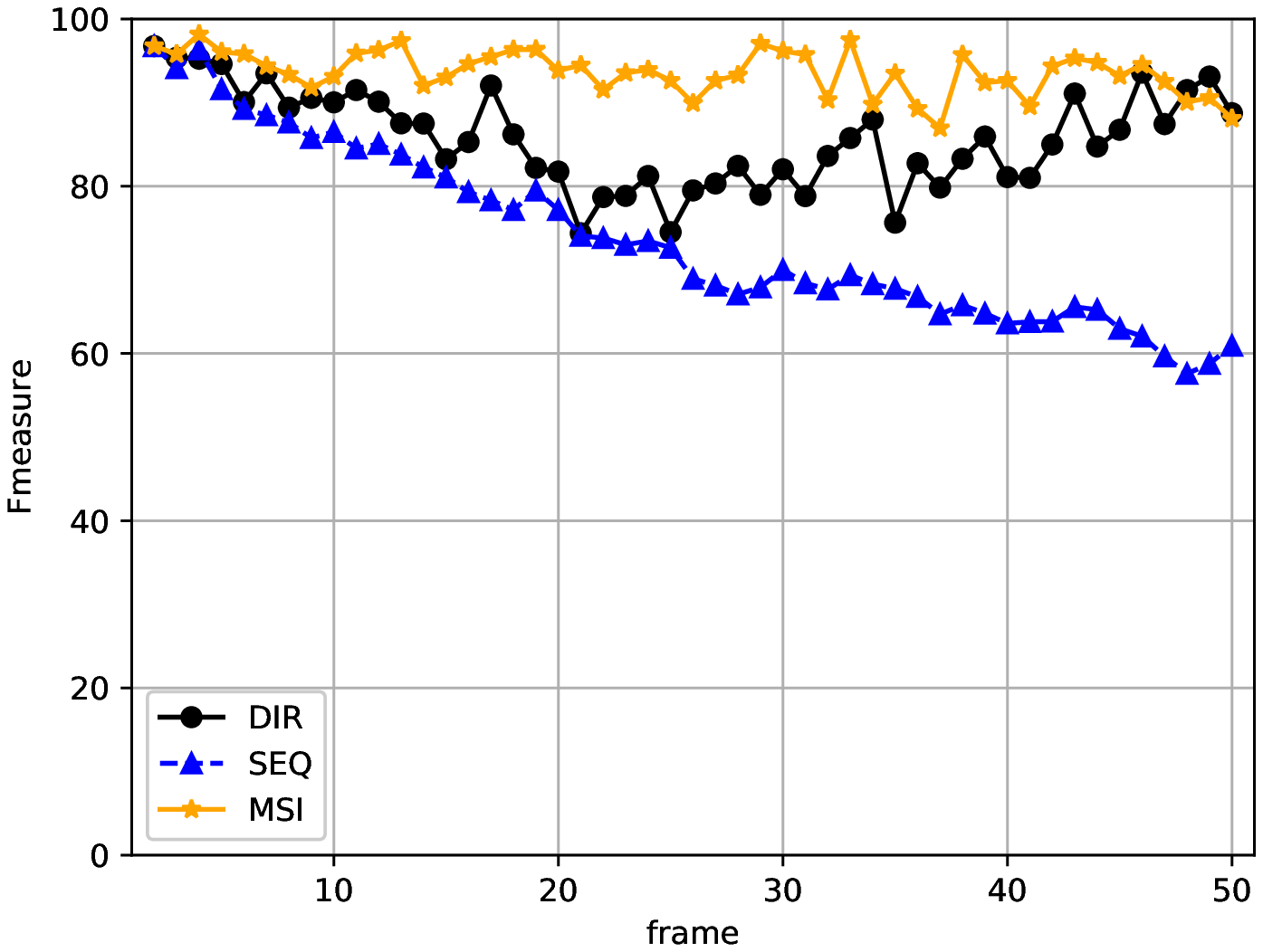} &
\hspace{-0.7cm} \includegraphics[width=4.3cm]{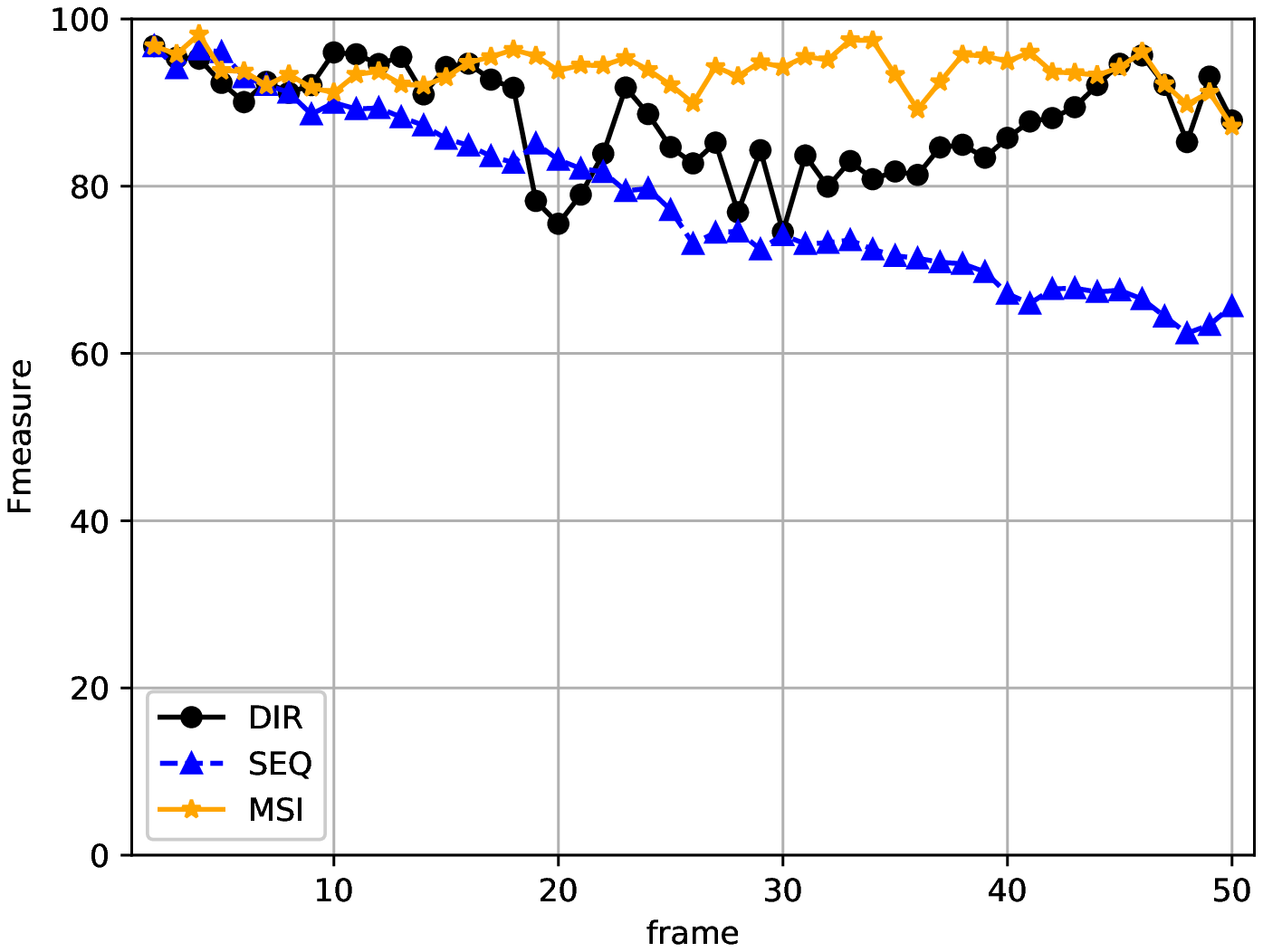} \vspace{-0.2cm}?\cr
\end{tabular}
\caption{Temporal evolution of DICE and F-measure scores during ROI tracking across \texttt{lapa} \cite{sznitman2012data}, \texttt{sle1.2} \cite{butler12eccv}, \texttt{octo} \cite{kristan2016vot} and \texttt{swan} \cite{perazzi2016benchmark} sequences. We compare direct (\texttt{DIR}), sequential (\texttt{SEQ}) and multi-step (\texttt{MSI}) integration based on $k$NN and RF-based elementary pairings.}
\label{fig::sec4-2-fig1}
\end{figure*}

Temporal evolutions of DICE and F-measure scores are displayed Fig.\ref{fig::sec4-2-fig1} along \texttt{lapa}, \texttt{sle1.2} (\texttt{sle1} for object $2$), \texttt{octo} and \texttt{swan} sequences with both classifiers. As already confirmed, best tracking results are reached with \texttt{MSI} compared to \texttt{DIR} and \texttt{SEQ}, especially for distant pairs. Contrary to \texttt{SEQ} whose performance decreases across sequences due to error accumulations (\texttt{lapa} and \texttt{swan} especially), multi-step estimations involved in \texttt{MSI} allow to fix uncorrect superpixel tracks as we can notice for \texttt{sle1.2} from frame $I_{20}$. Moreover, \texttt{DIR} is not prone to motion drift as \texttt{SEQ} but direct matching becomes tedious when inter-frame distances increase as shown for \texttt{octo} starting from frame $I_{109}$. Finally, it can be noticed that the temporal behavior remains the same regardless of the classifier.

\begin{figure}
\hspace{-2.05cm} \begin{minipage}{1\linewidth}
\begin{tabular}{ccccccc}
\hspace{-0.45cm} \includegraphics[width=2.585cm]{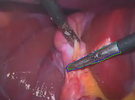} &
\hspace{-0.45cm} \includegraphics[width=2.585cm]{./figures/lapa/50-bound-SLIC.png} &
\hspace{-0.45cm} \includegraphics[width=2.585cm]{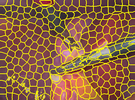} &
\hspace{-0.45cm} \includegraphics[width=2.585cm]{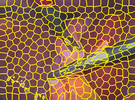} &
\hspace{-0.45cm} \includegraphics[width=2.585cm]{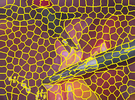} &
\hspace{-0.45cm} \includegraphics[width=2.585cm]{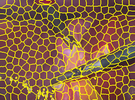} \vspace{-0.4cm} \cr
\hspace{-0.45cm} \tiny $I_{50}$ &
\hspace{-0.45cm} \tiny $I_{50}$ &
\hspace{-0.45cm} \tiny $I_{100}$ &
\hspace{-0.45cm} \tiny $I_{110}$ &
\hspace{-0.45cm} \tiny $I_{120}$ &
\hspace{-0.45cm} \tiny $I_{130}$ \cr
\end{tabular}
\begin{tabular}{ccccccc}
\hspace{-0.4cm} \rotatebox{90}{\tiny \textcolor{white}{-------} $I_{100}$} &
\hspace{-0.35cm} \includegraphics[width=2.5cm]{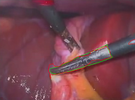} &
\hspace{-0.45cm} \includegraphics[width=2.5cm]{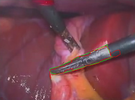} &
\hspace{-0.45cm} \includegraphics[width=2.5cm]{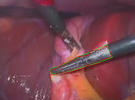} &
\hspace{-0.35cm} \includegraphics[width=2.5cm]{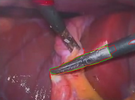} &
\hspace{-0.45cm} \includegraphics[width=2.5cm]{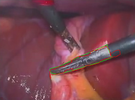} &
\hspace{-0.57cm} \includegraphics[width=2.5cm]{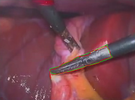} \cr
\hspace{-0.4cm} \rotatebox{90}{\tiny \textcolor{white}{-------} $I_{110}$} &
\hspace{-0.35cm} \includegraphics[width=2.5cm]{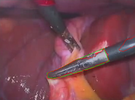} &
\hspace{-0.45cm} \includegraphics[width=2.5cm]{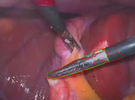} &
\hspace{-0.45cm} \includegraphics[width=2.5cm]{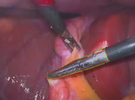} &
\hspace{-0.35cm} \includegraphics[width=2.5cm]{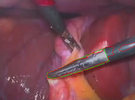} &
\hspace{-0.45cm} \includegraphics[width=2.5cm]{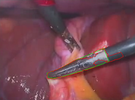} &
\hspace{-0.57cm} \includegraphics[width=2.5cm]{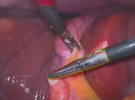} \cr
\hspace{-0.4cm} \rotatebox{90}{\tiny \textcolor{white}{-------} $I_{120}$} &
\hspace{-0.35cm} \includegraphics[width=2.5cm]{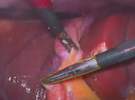} &
\hspace{-0.45cm} \includegraphics[width=2.5cm]{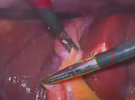} &
\hspace{-0.45cm} \includegraphics[width=2.5cm]{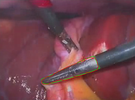} &
\hspace{-0.35cm} \includegraphics[width=2.5cm]{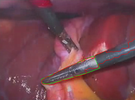} &
\hspace{-0.45cm} \includegraphics[width=2.5cm]{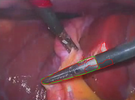} &
\hspace{-0.57cm} \includegraphics[width=2.5cm]{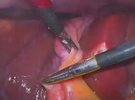} \cr
\hspace{-0.4cm} \rotatebox{90}{\tiny \textcolor{white}{-------} $I_{130}$} &
\hspace{-0.35cm} \includegraphics[width=2.5cm]{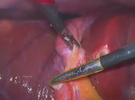} &
\hspace{-0.45cm} \includegraphics[width=2.5cm]{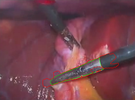} &
\hspace{-0.45cm} \includegraphics[width=2.5cm]{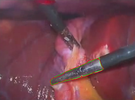} &
\hspace{-0.35cm} \includegraphics[width=2.5cm]{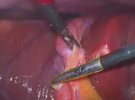} &
\hspace{-0.45cm} \includegraphics[width=2.5cm]{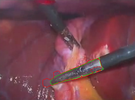} &
\hspace{-0.45cm} \includegraphics[width=2.5cm]{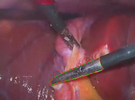} \vspace{-0.3cm} \cr
\hspace{-0.4cm} & \hspace{-0.35cm} \scriptsize $k$NN - \texttt{DIR} & \hspace{-0.45cm} \scriptsize $k$NN - \texttt{SEQ} & \hspace{-0.45cm} \scriptsize $k$NN - \texttt{MSI} & \hspace{-0.35cm} \scriptsize RF - \texttt{DIR} & \hspace{-0.45cm} \scriptsize RF - \texttt{SEQ} & \hspace{-0.57cm} \scriptsize RF - \texttt{MSI} \cr
\end{tabular} \end{minipage}
\caption{ROI selection and tracking across \texttt{lapa} sequence \cite{sznitman2012data} from $I_{50}$ to $I_{130}$. We compare direct (\texttt{DIR}), sequential (\texttt{SEQ}) and multi-step (\texttt{MSI}, steps $\{1,2,5,10,20,30,50\}$) integration (Sect.\ref{sec:sec3}) of superpixel matches obtained through unsupervised learning-based superpixel matching (Sect.\ref{sec:sec2}) with $k$NN and RF \cite{breiman2001random}. Superpixel decompositions are obtained \textit{via} SLIC \cite{achanta2012slic}. Blue boundaries ($I_{50}$) indicate superpixel labelling resulting from GT assignment. Green and red boundaries correspond resp. groundtruth (GT) and estimated tracking results.}
\label{fig::sec4-2-fig2}
\end{figure}

\begin{figure}
\hspace{-2.9cm} \begin{minipage}{1\linewidth}
\begin{tabular}{ccccccc}
\hspace{-0.45cm} \includegraphics[width=2.919cm]{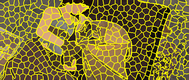} &
\hspace{-0.45cm} \includegraphics[width=2.919cm]{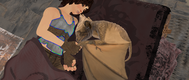} &
\hspace{-0.45cm} \includegraphics[width=2.919cm]{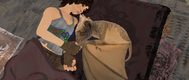} &
\hspace{-0.45cm} \includegraphics[width=2.919cm]{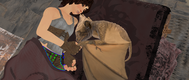} &
\hspace{-0.45cm} \includegraphics[width=2.919cm]{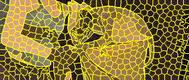} &
\hspace{-0.45cm} \includegraphics[width=2.919cm]{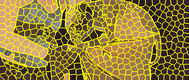} \vspace{-0.4cm} \cr
\hspace{-0.45cm} \tiny $I_{1}$ &
\hspace{-0.45cm} \tiny $I_{1}$ - \texttt{sle1.1} &
\hspace{-0.45cm} \tiny $I_{1}$ - \texttt{sle1.2} &
\hspace{-0.45cm} \tiny $I_{1}$ - \texttt{sle1.3} &
\hspace{-0.45cm} \tiny $I_{40}$ &
\hspace{-0.45cm} \tiny $I_{50}$ \cr
\end{tabular}
\begin{tabular}{cccccccc}
\hspace{-0.4cm} \rotatebox{90}{\tiny \textcolor{white}{--} \texttt{sle1.1}} &
\hspace{-0.4cm} \rotatebox{90}{\tiny \textcolor{white}{---} $I_{40}$} &
\hspace{-0.35cm} \includegraphics[width=2.8cm]{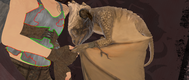} &
\hspace{-0.45cm} \includegraphics[width=2.8cm]{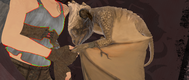} &
\hspace{-0.45cm} \includegraphics[width=2.8cm]{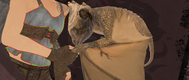} &
\hspace{-0.4cm} \includegraphics[width=2.8cm]{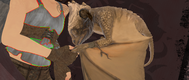} &
\hspace{-0.45cm} \includegraphics[width=2.8cm]{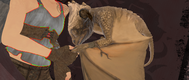} &
\hspace{-0.56cm} \includegraphics[width=2.8cm]{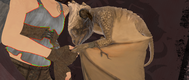} \cr
\hspace{-0.4cm} \rotatebox{90}{\tiny \textcolor{white}{--} \texttt{sle1.1}} &
\hspace{-0.4cm} \rotatebox{90}{\tiny \textcolor{white}{---} $I_{50}$} &
\hspace{-0.35cm} \includegraphics[width=2.8cm]{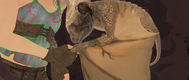} &
\hspace{-0.45cm} \includegraphics[width=2.8cm]{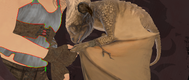} &
\hspace{-0.45cm} \includegraphics[width=2.8cm]{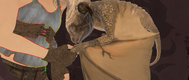} &
\hspace{-0.4cm} \includegraphics[width=2.8cm]{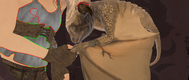} &
\hspace{-0.45cm} \includegraphics[width=2.8cm]{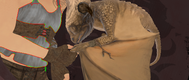} &
\hspace{-0.56cm} \includegraphics[width=2.8cm]{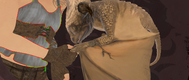} \cr
\hspace{-0.4cm} \rotatebox{90}{\tiny \textcolor{white}{--} \texttt{sle1.2}} &
\hspace{-0.4cm} \rotatebox{90}{\tiny \textcolor{white}{---} $I_{40}$} &
\hspace{-0.35cm} \includegraphics[width=2.8cm]{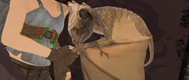} &
\hspace{-0.45cm} \includegraphics[width=2.8cm]{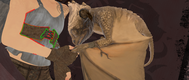} &
\hspace{-0.45cm} \includegraphics[width=2.8cm]{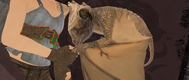} &
\hspace{-0.4cm} \includegraphics[width=2.8cm]{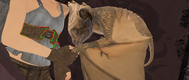} &
\hspace{-0.45cm} \includegraphics[width=2.8cm]{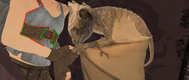} &
\hspace{-0.56cm} \includegraphics[width=2.8cm]{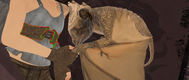} \cr
\hspace{-0.4cm} \rotatebox{90}{\tiny \textcolor{white}{--} \texttt{sle1.1}} &
\hspace{-0.4cm} \rotatebox{90}{\tiny \textcolor{white}{---} $I_{50}$} &
\hspace{-0.35cm} \includegraphics[width=2.8cm]{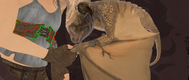} &
\hspace{-0.45cm} \includegraphics[width=2.8cm]{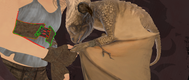} &
\hspace{-0.45cm} \includegraphics[width=2.8cm]{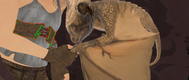} &
\hspace{-0.4cm} \includegraphics[width=2.8cm]{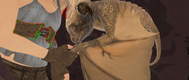} &
\hspace{-0.45cm} \includegraphics[width=2.8cm]{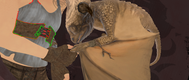} &
\hspace{-0.56cm} \includegraphics[width=2.8cm]{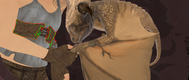} \cr
\hspace{-0.4cm} \rotatebox{90}{\tiny \textcolor{white}{--} \texttt{sle1.3}} &
\hspace{-0.4cm} \rotatebox{90}{\tiny \textcolor{white}{---} $I_{40}$} &
\hspace{-0.35cm} \includegraphics[width=2.8cm]{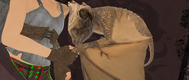} &
\hspace{-0.45cm} \includegraphics[width=2.8cm]{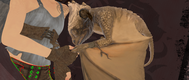} &
\hspace{-0.45cm} \includegraphics[width=2.8cm]{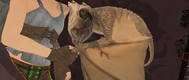} &
\hspace{-0.4cm} \includegraphics[width=2.8cm]{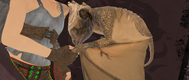} &
\hspace{-0.45cm} \includegraphics[width=2.8cm]{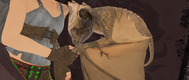} &
\hspace{-0.56cm} \includegraphics[width=2.8cm]{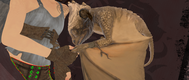} \cr
\hspace{-0.4cm} \rotatebox{90}{\tiny \textcolor{white}{--} \texttt{sle1.1}} &
\hspace{-0.4cm} \rotatebox{90}{\tiny \textcolor{white}{---} $I_{50}$} &
\hspace{-0.35cm} \includegraphics[width=2.8cm]{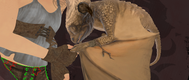} &
\hspace{-0.45cm} \includegraphics[width=2.8cm]{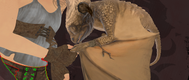} &
\hspace{-0.45cm} \includegraphics[width=2.8cm]{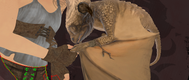} &
\hspace{-0.4cm} \includegraphics[width=2.8cm]{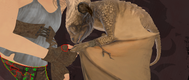} &
\hspace{-0.45cm} \includegraphics[width=2.8cm]{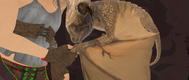} &
\hspace{-0.45cm} \includegraphics[width=2.8cm]{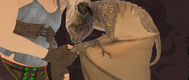} \vspace{-0.3cm} \cr
\hspace{-0.4cm} & 
\hspace{-0.4cm} & 
\hspace{-0.35cm} \scriptsize $k$NN - \texttt{DIR} &
\hspace{-0.45cm} \scriptsize $k$NN - \texttt{SEQ} & 
\hspace{-0.45cm} \scriptsize $k$NN - \texttt{MSI} & 
\hspace{-0.4cm}\scriptsize RF - \texttt{DIR} & 
\hspace{-0.45cm}\scriptsize RF - \texttt{SEQ} &
\hspace{-0.56cm} \scriptsize RF - \texttt{MSI} \cr
\end{tabular} \end{minipage} \vspace{-0.2cm}
\caption{ROI selections and tracking across \texttt{sle1} sequence \cite{butler12eccv} from $I_{1}$ to $I_{50}$. We compare direct (\texttt{DIR}), sequential (\texttt{SEQ}) and multi-step (\texttt{MSI}, steps $\{1,2,5,10,20\}$) integration based on unsupervised learning-based superpixel matching with $k$NN and RF.}
\label{fig::sec4-2-fig4}
\end{figure}

\begin{figure}
\centering \begin{tabular}{ccccccc}
\hspace{-0.45cm} \includegraphics[width=1.714cm]{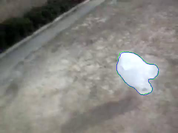} &
\hspace{-0.45cm} \includegraphics[width=1.714cm]{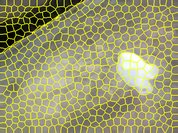} &
\hspace{-0.45cm} \includegraphics[width=1.714cm]{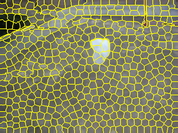} &
\hspace{-0.45cm} \includegraphics[width=1.714cm]{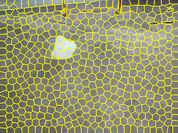} &
\hspace{-0.45cm} \includegraphics[width=1.714cm]{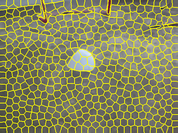} \vspace{-0.4cm} \cr
\hspace{-0.45cm} \tiny $I_{1}$ &
\hspace{-0.45cm} \tiny $I_{1}$ &
\hspace{-0.45cm} \tiny $I_{51}$ &
\hspace{-0.45cm} \tiny $I_{71}$ &
\hspace{-0.45cm} \tiny $I_{101}$ \cr
\end{tabular} \vspace{-0.1cm} \\
\begin{tabular}{ccccccc}
\hspace{-0.45cm} \includegraphics[width=1.714cm]{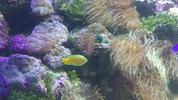} &
\hspace{-0.45cm} \includegraphics[width=1.714cm]{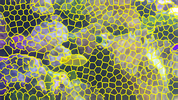} &
\hspace{-0.45cm} \includegraphics[width=1.714cm]{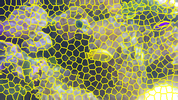} &
\hspace{-0.45cm} \includegraphics[width=1.714cm]{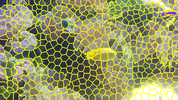} &
\hspace{-0.45cm} \includegraphics[width=1.714cm]{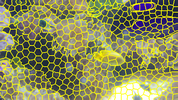} \vspace{-0.4cm} \cr
\hspace{-0.45cm} \tiny $I_{301}$ &
\hspace{-0.45cm} \tiny $I_{301}$ &
\hspace{-0.45cm} \tiny $I_{351}$ &
\hspace{-0.45cm} \tiny $I_{371}$ &
\hspace{-0.45cm} \tiny $I_{401}$ \cr
\end{tabular} \vspace{-0.1cm} \\
\begin{tabular}{ccccccc}
\hspace{-0.45cm} \includegraphics[width=1.714cm]{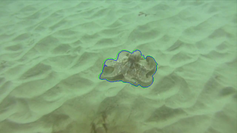} &
\hspace{-0.45cm} \includegraphics[width=1.714cm]{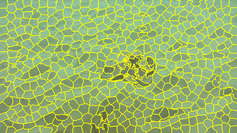} &
\hspace{-0.45cm} \includegraphics[width=1.714cm]{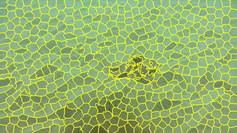} &
\hspace{-0.45cm} \includegraphics[width=1.714cm]{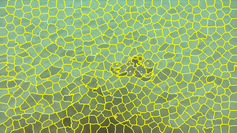} &
\hspace{-0.45cm} \includegraphics[width=1.714cm]{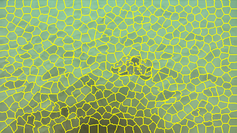} \vspace{-0.4cm} \cr
\hspace{-0.45cm} \tiny $I_{70}$ &
\hspace{-0.45cm} \tiny $I_{70}$ &
\hspace{-0.45cm} \tiny $I_{90}$ &
\hspace{-0.45cm} \tiny $I_{110}$ &
\hspace{-0.45cm} \tiny $I_{120}$ \cr
\end{tabular}
\begin{tabular}{ccccc}
\hspace{-0.5cm} \rotatebox{90}{\tiny \textcolor{white}{------} $I_{51}$} &
\hspace{-0.4cm} \includegraphics[width=2.1cm]{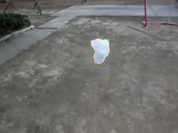} &
\hspace{-0.45cm} \includegraphics[width=2.1cm]{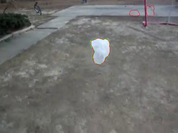} &
\hspace{-0.45cm} \includegraphics[width=2.1cm]{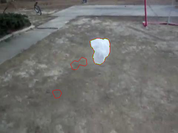} &
\hspace{-0.58cm} \includegraphics[width=2.1cm]{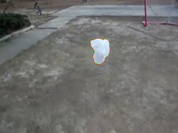} \cr
\hspace{-0.5cm} \rotatebox{90}{\tiny \textcolor{white}{------} $I_{71}$} &
\hspace{-0.4cm} \includegraphics[width=2.1cm]{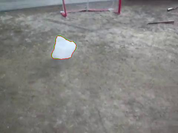} &
\hspace{-0.45cm} \includegraphics[width=2.1cm]{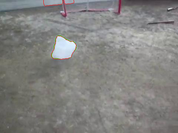} &
\hspace{-0.45cm} \includegraphics[width=2.1cm]{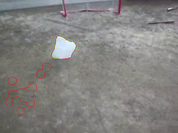} &
\hspace{-0.58cm} \includegraphics[width=2.1cm]{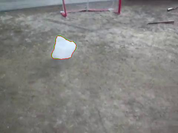} \cr
\hspace{-0.5cm} \rotatebox{90}{\tiny \textcolor{white}{------} $I_{101}$} &
\hspace{-0.4cm} \includegraphics[width=2.1cm]{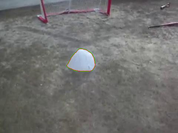} &
\hspace{-0.45cm} \includegraphics[width=2.1cm]{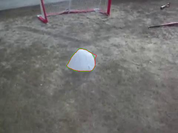} &
\hspace{-0.45cm} \includegraphics[width=2.1cm]{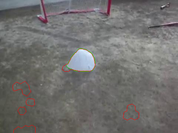} &
\hspace{-0.45cm} \includegraphics[width=2.1cm]{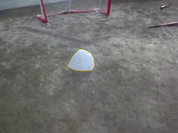} \vspace{-0.3cm} \cr
\end{tabular} \vspace{0.235cm} \\
\begin{tabular}{ccccc}
\hspace{-0.5cm} \rotatebox{90}{\tiny \textcolor{white}{---} $I_{351}$} &
\hspace{-0.4cm} \includegraphics[width=2.1cm]{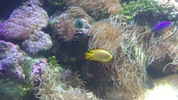} &
\hspace{-0.45cm} \includegraphics[width=2.1cm]{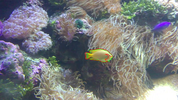} &
\hspace{-0.45cm} \includegraphics[width=2.1cm]{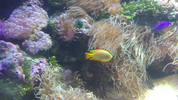} &
\hspace{-0.58cm} \includegraphics[width=2.1cm]{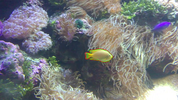} \cr
\hspace{-0.5cm} \rotatebox{90}{\tiny \textcolor{white}{---} $I_{371}$} &
\hspace{-0.4cm} \includegraphics[width=2.1cm]{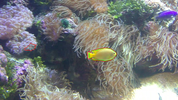} &
\hspace{-0.45cm} \includegraphics[width=2.1cm]{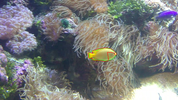} &
\hspace{-0.45cm} \includegraphics[width=2.1cm]{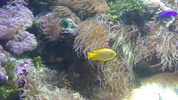} &
\hspace{-0.58cm} \includegraphics[width=2.1cm]{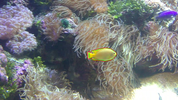} \cr
\hspace{-0.5cm} \rotatebox{90}{\tiny \textcolor{white}{---} $I_{401}$} &
\hspace{-0.4cm} \includegraphics[width=2.1cm]{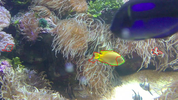} &
\hspace{-0.45cm} \includegraphics[width=2.1cm]{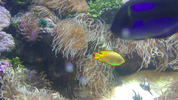} &
\hspace{-0.45cm} \includegraphics[width=2.1cm]{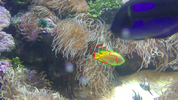} &
\hspace{-0.45cm} \includegraphics[width=2.1cm]{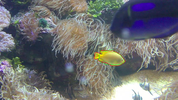} \vspace{-0.3cm} \cr
\end{tabular} \vspace{0.235cm} \\
\begin{tabular}{ccccc}
\hspace{-0.5cm} \rotatebox{90}{\tiny \textcolor{white}{---} $I_{90}$} &
\hspace{-0.4cm} \includegraphics[width=2.1cm]{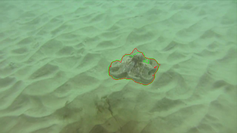} &
\hspace{-0.45cm} \includegraphics[width=2.1cm]{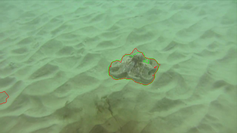} &
\hspace{-0.45cm} \includegraphics[width=2.1cm]{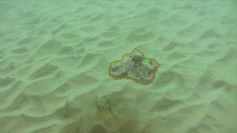} &
\hspace{-0.58cm} \includegraphics[width=2.1cm]{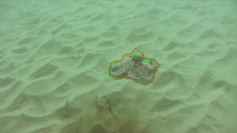} \cr
\hspace{-0.5cm} \rotatebox{90}{\tiny \textcolor{white}{---} $I_{110}$} &
\hspace{-0.4cm} \includegraphics[width=2.1cm]{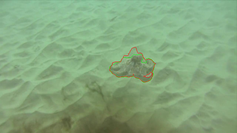} &
\hspace{-0.45cm} \includegraphics[width=2.1cm]{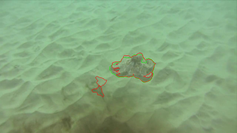} &
\hspace{-0.45cm} \includegraphics[width=2.1cm]{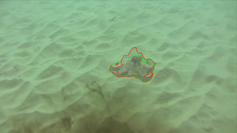} &
\hspace{-0.58cm} \includegraphics[width=2.1cm]{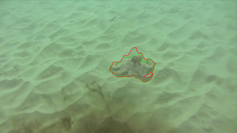} \cr
\hspace{-0.5cm} \rotatebox{90}{\tiny \textcolor{white}{---} $I_{120}$} &
\hspace{-0.4cm} \includegraphics[width=2.1cm]{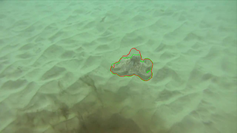} &
\hspace{-0.45cm} \includegraphics[width=2.1cm]{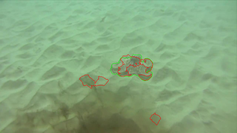} &
\hspace{-0.45cm} \includegraphics[width=2.1cm]{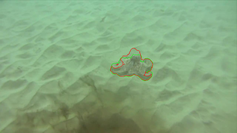} &
\hspace{-0.45cm} \includegraphics[width=2.1cm]{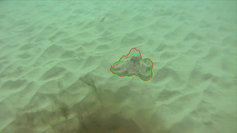}
\vspace{-0.3cm} \cr
\hspace{-0.55cm} & 
\hspace{-0.4cm} \scriptsize $k$NN - \texttt{MSI} &
\hspace{-0.45cm} \scriptsize RF - \texttt{DIR} &
\hspace{-0.45cm} \scriptsize RF - \texttt{SEQ} &
\hspace{-0.45cm} \scriptsize RF - \texttt{MSI} \cr
\end{tabular} \vspace{-0.2cm} 

\caption{ROI selection and tracking across \texttt{bag} (resp. \texttt{fsh3}, \texttt{octo}) sequences \cite{kristan2016vot} for segments $[I_{1},I_{101}]$ ($[I_{301},I_{401}]$, $[I_{70},I_{120}]$). We compare \texttt{DIR}, \texttt{SEQ} and \texttt{MSI} (steps $\{1,2,5,10,20,30,50\}$ for \texttt{bag} and \texttt{fsh3}, $\{1,2,5,10,20\}$ for \texttt{octo}) integration based on unsupervised learning-based superpixel matching with $k$NN and RF \cite{breiman2001random}.}
\label{fig::sec4-2-fig3} 
\end{figure}

Finally, quantitative results are illustrated by series of ROI selection and visual tracking examples for several pairs of \texttt{lapa} (Fig.\ref{fig::sec4-2-fig2}), \texttt{sle1} (Fig.\ref{fig::sec4-2-fig4}), \texttt{bag}, \texttt{fsh3}, and \texttt{octo} (Fig.\ref{fig::sec4-2-fig3}) sequences. Fig.\ref{fig::sec4-2-fig2} shows that $k$NN-based \texttt{MSI} provides a very good delineation of the surgical tool for all image pairs, which suggests that series of medical images can be also processed with the proposed methodology. Reliable ROI tracking through \texttt{MSI} is also shown on synthetic images (Fig.\ref{fig::sec4-2-fig4}) despite strong scale variations. Propagation of matching errors with \texttt{SEQ} is clearly illustrated Fig.\ref{fig::sec4-2-fig2} for \texttt{lapa} ($k$NN) and Fig.\ref{fig::sec4-2-fig3} for \texttt{bag}. Tracking failures with \texttt{DIR} are temporally uncorrelated but strong enough to damage the propagation task for the fish (Fig.\ref{fig::sec4-2-fig3}) due to color variations of its right part. ROI tracking for \texttt{octo} (Fig.\ref{fig::sec4-2-fig3}) gives correct results both with $k$NN and RF despite significant color similarities with the dynamic background. Results provided Fig.\ref{fig::sec4-1-fig1},\ref{fig::sec4-2-fig4} demonstrate that RF-based \texttt{MSI} outperforms $k$NN-based \texttt{MSI} as well as RF-based \texttt{DIR} and \texttt{SEQ} for \texttt{swan} and \texttt{sle1} videos. Another visualization through prediction masks given Fig.\ref{fig::sec4-1-fig1} for the \texttt{swan} pair confirms the ability of RF-based \texttt{MSI} to reach accurate long-term correspondences (see for instance the swan beak). Indeed, a given object part must keep the same color between training and prediction in case of correct matching.

\subsection{Long-term candidate generation}
\label{ssec:sec4-3}

We propose to perform a more in-depth study of multi-step integration by comparing different long-term candidate generation strategies in terms of tracking accuracy. As described Sect.\ref{ssec:sec3-3}, long-term superpixel candidates can be generated using direct candidates only (\texttt{MSId}), both direct and reverse candidates (\texttt{MSIr}) or only candidates generated in both direct and reverse directions (\texttt{MSIm}). Note that the previously given \texttt{MSI} results corresponded to \texttt{MSIm} where only superpixel duplicates are taken into account for majority voting (Eq.\ref{eq::sec3-3-eq1}). \texttt{MSId}, \texttt{MSIr} and \texttt{MSIm} are comparatively evaluated in terms of tracking accuracy based on RF-based superpixel elementary matches. Results are provided Tab.\ref{tab::sec4-3-tab1} through DICE, F-measure and consistency scores across the $10$ sequences used for ROI tracking. 

\begin{table}
\scriptsize
\centering \hspace{-0.1cm}\begin{tabular}{l?c|c|c?c|c|c?c|c|c|}
\cline{2-10}
\multirow{2}{*}{} & \multicolumn{3}{c?}{DICE} & \multicolumn{3}{c?}{F-measure} & \multicolumn{3}{c|}{consistency}  \cr \cline{2-10}
& \texttt{MSId} & \texttt{MSIr} & \texttt{MSIm} & 
\texttt{MSId} & \texttt{MSIr} & \texttt{MSIm} & 
\texttt{MSId} & \texttt{MSIr} & \texttt{MSIm} \cr \hline 
\multicolumn{1}{|l?}{\hspace{-0.2cm} \texttt{bag} \hspace{-0.2cm} } & 84.5 & 91.9 & \textbf{92.9} &
80.3 & \textbf{88.3} & \textbf{88.3} &
19.2 & \textbf{27.8} & \textbf{27.8} \cr \hline
\multicolumn{1}{|l?}{\hspace{-0.2cm} \texttt{fsh3} \hspace{-0.2cm} } & \textbf{92.0} & \textbf{92.0} & \textbf{92.0} &
94.9 & \textbf{95.4} & \textbf{95.4} &
52.8 & \textbf{59.6} & 58.7 \cr \hline
\multicolumn{1}{|l?}{\hspace{-0.2cm} \texttt{octo} \hspace{-0.2cm} } & 91.8 & 92.2 & \textbf{92.8} &
83.8 & 84.4 & \textbf{86.0} &
71.6 & \textbf{75.8} & 75.7 \cr \hline
\multicolumn{1}{|l?}{\hspace{-0.2cm} \texttt{lapa} \hspace{-0.2cm} } & 92.6 & \textbf{92.9} & 92.8 &
94.2 & \textbf{95.2} & 94.8 &
85.4 & \textbf{91.5} & \textbf{91.5} \cr \hline
\multicolumn{1}{|l?}{\hspace{-0.2cm} \texttt{sl1.1} \hspace{-0.2cm} } & 95.7 & 95.8 & \textbf{95.9} &
94.5 & 94.8 & \textbf{95.5} &
91.2 & \textbf{93.7} & 93.4 \cr \hline
\multicolumn{1}{|l?}{\hspace{-0.2cm} \texttt{sl1.2} \hspace{-0.2cm} } & 90.3 & \textbf{91.0} & 90.9 &
91.8 & \textbf{94.5} & 93.9 &
78.6 & 90.4 & \textbf{91.2} \cr \hline
\multicolumn{1}{|l?}{\hspace{-0.2cm} \texttt{sl1.3} \hspace{-0.2cm} } & 94.2 & \textbf{94.3} & 94.2 &
81.3 & 81.5 & \textbf{81.6} &
90.0 & 99.5 & \textbf{100} \cr \hline
\multicolumn{1}{|l?}{\hspace{-0.2cm} \texttt{swan} \hspace{-0.2cm} } & 93.2 & 93.1 & \textbf{93.4} &
93.4 & 93.0 & \textbf{93.8} &
70.6 & 77.6 & \textbf{77.7} \cr \hline
\multicolumn{1}{|l?}{\hspace{-0.2cm} \texttt{bear} \hspace{-0.2cm} } & 92.2 & 92.5 & \textbf{92.5} &
\textbf{83.3} & 82.9 & 82.1 &
54.7 & 67.0 & \textbf{67.2} \cr \hline
\multicolumn{1}{|l?}{\hspace{-0.2cm} \texttt{caml} \hspace{-0.2cm} } & \textbf{79.6} & 79.3 & \textbf{79.6} &
\textbf{69.6} & 69.3 & 69.1 &
44.2 & 54.1 & \textbf{54.7} \cr \hline
\multicolumn{1}{|l?}{\hspace{-0.2cm} \texttt{cows} \hspace{-0.2cm} } & \textbf{90.0} & 89.3 & 89.1 &
\textbf{75.5} & 74.4 & 74.1 &
59.8 & 67.2 & \textbf{67.6} \cr \hline
\multicolumn{1}{|l?}{\hspace{-0.2cm} \texttt{flam} \hspace{-0.2cm} } & 79.8 & 80.7 & \textbf{80.8} &
64.0 & 69.0 & \textbf{71.9} &
40.9 & 54.3 & \textbf{55.0} \cr \hline \hline
\multicolumn{1}{|l?}{\hspace{-0.2cm} \textit{avg} \hspace{-0.2cm} } & \textit{89.7} & \textit{90.4} & \textit{\textbf{90.6}} &
\textit{83.9} & \textit{85.2} & \textit{\textbf{85.5}} &
\textit{63.3} & \textit{71.5} & \textit{\textbf{71.7}} \cr \hline
\end{tabular} \vspace{0.1cm}
\caption{DICE, F-measure and consistency scores for ROI tracking across $10$ sequences. Based on RF-based superpixel elementary matches, we compare three different superpixel candidate generation strategies for multi-step (\texttt{MSI}) integration using: only direct candidates (\texttt{MSId}), direct and reverse candidates (\texttt{MSIr}), only candidates generated in both direct and reverse directions (\texttt{MSIm}). Bold results indicate the best performance.}
\label{tab::sec4-3-tab1}
\end{table}

Results from Tab.\ref{tab::sec4-3-tab1} bring two main findings. First, we observe that tracking accuracy is improved when reverse candidates are used additionally to direct ones. Consistency ratios are clearly improved (from $63.3$ to $71.5\%$ when comparing \texttt{MSId}/\texttt{MSIr}) as expected but DICE and F-measure improvements can be also observed with gains of $0.7$ and $1.3$ between \texttt{MSId}/\texttt{MSIr}. Second, relying on superpixel duplicates only (\texttt{MSIm}) brings a slighltly better ROI tracking compared to \texttt{MSIr}. Average results slightly increase from $90.4$ to $90.6$, $85.2$ to $85.5$ and $71.5$ to $71.7$ for  DICE, F-measure and consistency which shows that extensive mutual matching guidance as performed with \texttt{MSIm} is the best strategy to perform long-term superpixel tracking from multi-step elementary correspondences.

\section{Conclusion}
\label{sec:sec5}

In this work, we proposed a two-step pipeline dedicated to long-term superpixel tracking. unsupervised learning-based superpixel matching is firstly considered as an elementary displacement estimator to provide correspondences between consecutive and distant images using either nearest neighbors or random forests with robust context-rich features we extended from greyscale to multi-channel and forward-backward consistency contraints. Resulting elementary matches are then combined along multi-step paths running through the sequence with various inter-frame distances to produce a large set of candidate long-term superpixel pairings upon which majority voting selection is performed. Compared to state-of-the-art methods including pixel or patch-based strategies which may suffer from regular support regions, video object tracking experiments demonstrate that unsupervised learning can produce reliable correspondences between semantically meaningful areas. Moreover, the ability of multi-step integration to combine these pairings towards accurate long-term superpixel tracking has been shown against usual direct and sequential integrations. Extending this work from single to hierarchical multi-scale superpixel decomposition would deserve further investigation for future research since dealing with multiple spatial extends can drive the matching process in a coarse-to-fine fashion. Other features such as spectral features could be employed to further improve unsupervised learning-based matching while reducing processing time. In addition, very long-term superpixel tracking could be reached by combining superpixel pairings estimated with respect to multiple reference frames. Our contributions also give new insights for optical flow and registration initialization, in particular to provide a better management of  large displacements, appearance and illumination changes. More generally, the proposed framework could be easily extended to other imaging modalities including series of medical images for anatomical structure tracking.


\bibliography{conze-IVC-2018}

\begin{thebibliography}{10}
\expandafter\ifx\csname url\endcsname\relax
  \def\url#1{\texttt{#1}}\fi
\expandafter\ifx\csname urlprefix\endcsname\relax\def\urlprefix{URL }\fi
\expandafter\ifx\csname href\endcsname\relax
  \def\href#1#2{#2} \def\path#1{#1}\fi

\bibitem{lezama2011track}
J.~Lezama, K.~Alahari, J.~Sivic, I.~Laptev, Track to the future:
  Spatio-temporal video segmentation with long-range motion cues, in: IEEE
  Conference on Computer Vision and Pattern Recognition, 2011, pp. 3369--3376.

\bibitem{seitz2006comparison}
S.~M. Seitz, B.~Curless, J.~Diebel, D.~Scharstein, R.~Szeliski, A comparison
  and evaluation of multi-view stereo reconstruction algorithms, in: IEEE
  Conference on Computer Vision and Pattern Recognition, 2006, pp. 519--528.

\bibitem{yang2014robust}
F.~Yang, H.~Lu, M.-H. Yang, Robust superpixel tracking, IEEE Transactions on
  Image Processing 23~(4) (2014) 1639--1651.

\bibitem{wang2013dense}
H.~Wang, A.~Kl{\"a}ser, C.~Schmid, C.-L. Liu, Dense trajectories and motion
  boundary descriptors for action recognition, International Journal of
  Computer Vision 103~(1) (2013) 60--79.

\bibitem{cao2011semi}
X.~Cao, Z.~Li, Q.~Dai, Semi-automatic 2{D}-to-3{D} conversion using disparity
  propagation, IEEE Transactions on Broadcasting 57~(2) (2011) 491--499.

\bibitem{conze2016multi}
P.-H. Conze, P.~Robert, T.~Crivelli, L.~Morin, Multi-reference combinatorial
  strategy towards longer long-term dense motion estimation, Computer Vision
  and Image Understanding 150 (2016) 66--80.

\bibitem{farneback2003two}
G.~Farneb{\"a}ck, Two-frame motion estimation based on polynomial expansion,
  Image Analysis (2003) 363--370.

\bibitem{shi1994good}
J.~Shi, C.~Tomasi, Good features to track, in: IEEE International Conference on
  Computer Vision and Pattern Recognition, 1994, pp. 593--600.

\bibitem{barnes2009patchmatch}
C.~Barnes, E.~Shechtman, A.~Finkelstein, D.~B. Goldman, Patchmatch: A
  randomized correspondence algorithm for structural image editing, ACM
  Transaction on Graphics 28~(3) (2009) 24--1.

\bibitem{barnes2010generalized}
C.~Barnes, E.~Shechtman, D.~Goldman, A.~Finkelstein, The generalized patchmatch
  correspondence algorithm, European Conference on Computer Vision (2010)
  29--43.

\bibitem{achanta2012slic}
R.~Achanta, A.~Shaji, K.~Smith, A.~Lucchi, P.~Fua, S.~Susstrunk, {SLIC}
  superpixels compared to state-of-the-art superpixel methods, IEEE
  Transactions on Pattern Analysis and Machine Intelligence 34 (2012)
  2274--2282.

\bibitem{chang2013superpixel}
H.-S. Chang, Y.-C.~F. Wang, Superpixel-based large displacement optical flow,
  in: IEEE International Conference on Image Processing, 2013, pp. 3835--3839.

\bibitem{donne2015fast}
S.~Donn{\'e}, J.~Aelterman, B.~Goossens, W.~Philips, Fast and robust
  variational optical flow for high-resolution images using {SLIC} superpixels,
  in: International Conference on Advanced Concepts for Intelligent Vision
  Systems, 2015, pp. 205--216.

\bibitem{dong2016hsp2p}
X.~Dong, J.~Shen, L.~Shao, {HSP2P}: Hierarchical superpixel-to-pixel dense
  image matching, IEEE Transactions on Circuits and Systems for Video
  Technology --~(--) (2017) --, not fully edited.

\bibitem{fan2016visual}
H.~Fan, J.~Xiang, Z.~Chen, Visual tracking by local superpixel matching with
  markov random field, in: Pacific Rim Conference on Multimedia, 2016.

\bibitem{giraud2017superpatchmatch}
R.~Giraud, V.~T. Ta, A.~Bugeau, P.~Coupé, N.~Papadakis, Superpatchmatch: An
  algorithm for robust correspondences using superpixel patches, IEEE
  Transactions on Image Processing 26~(8) (2017) 4068--4078.

\bibitem{kanavati2017supervoxel}
F.~Kanavati, T.~Tong, K.~Misawa, M.~Fujiwara, K.~Mori, D.~Rueckert, B.~Glocker,
  Supervoxel classification forests for estimating pairwise image
  correspondences, Pattern Recognition 63 (2017) 561--569.

\bibitem{breiman2001random}
L.~Breiman, Random {F}orests, Machine learning 45~(1) (2001) 5--32.

\bibitem{crivelli2015robust}
T.~Crivelli, M.~Fradet, P.-H. Conze, P.~Robert, P.~P\'erez, Robust optical flow
  integration, IEEE Transactions on Image Processing 24~(1) (2015) 484--498.

\bibitem{wang2017constrained}
L.~Wang, H.~Lu, M.~H. Yang, Constrained superpixel tracking, IEEE Transactions
  on Cybernetics~(99) (2017) 1--12.

\bibitem{ma2015hierarchical}
C.~Ma, J.-B. Huang, X.~Yang, M.-H. Yang, Hierarchical convolutional features
  for visual tracking, in: IEEE International Conference on Computer Vision,
  2015, pp. 3074--3082.

\bibitem{li2016deeptrack}
H.~Li, Y.~Li, F.~Porikli, Deeptrack: Learning discriminative feature
  representations online for robust visual tracking, IEEE Transactions on Image
  Processing 25~(4) (2016) 1834--1848.

\bibitem{zhang2017deep}
D.~Zhang, H.~Maei, X.~Wang, Y.-F. Wang, Deep reinforcement learning for visual
  object tracking in videos, arXiv preprint 1701.08936.

\bibitem{glocker2014robust}
B.~Glocker, D.~Zikic, D.~R. Haynor, Robust registration of longitudinal spine
  {CT}, in: Medical Image Computing and Computer-Assisted Intervention, 2014,
  pp. 251--258.

\bibitem{conze2017semi}
P.-H. Conze, F.~Tilquin, V.~Noblet, F.~Rousseau, F.~Heitz, P.~Pessaux,
  Hierarchical multi-scale supervoxel matching using random forests for
  automatic semi-dense abdominal image registration, in: IEEE International
  Symposium on Biomedical Imaging, 2017, pp. 490--493.

\bibitem{sznitman2012data}
R.~Sznitman, K.~Ali, R.~Richa, R.~H. Taylor, G.~D. Hager, P.~Fua, Data-driven
  visual tracking in retinal microsurgery, in: International Conference on
  Medical Image Computing and Computer-Assisted Intervention, Springer, 2012,
  pp. 568--575.

\bibitem{brox2011large}
T.~Brox, J.~Malik, Large displacement optical flow: descriptor matching in
  variational motion estimation, IEEE Transactions on Pattern Analysis and
  Machine Intelligence 33~(3) (2011) 500--513.

\bibitem{roth2009discrete}
S.~Roth, V.~Lempitsky, C.~Rother, Discrete-continuous optimization for optical
  flow estimation, in: Statistical and Geometrical Approaches to Visual Motion
  Analysis, 2009, pp. 1--22.

\bibitem{kristan2016vot}
M.~Kristan, {et al.}, The visual object tracking {VOT}2016 challenge results,
  in: Workshop of European Conference on Computer Vision, 2016, pp. 777--823.

\bibitem{butler12eccv}
D.~J. Butler, J.~Wulff, G.~B. Stanley, M.~J. Black, A naturalistic open source
  movie for optical flow evaluation, in: European Conference on Computer
  Vision, 2012, pp. 611--625.

\bibitem{perazzi2016benchmark}
F.~Perazzi, J.~Pont-Tuset, B.~McWilliams, L.~Van~Gool, M.~Gross,
  A.~Sorkine-Hornung, A benchmark dataset and evaluation methodology for video
  object segmentation, in: IEEE Conference on Computer Vision and Pattern
  Recognition, 2016, pp. 724--732.

\bibitem{martin2004learning}
D.~R. Martin, C.~C. Fowlkes, J.~Malik, Learning to detect natural image
  boundaries using local brightness, color, and texture cues, IEEE Transactions
  on Pattern Analysis and Machine Intelligence 26~(5) (2004) 530--549.

\bibitem{liu2011sift}
C.~Liu, J.~Yuen, A.~Torralba, Sift flow: Dense correspondence across scenes and
  its applications, IEEE Transactions on Pattern Analysis and Machine
  Intelligence 33~(5) (2011) 978--994.

\end{thebibliography}

\end{document}